\documentclass{article}

\usepackage{microtype}
\usepackage{graphicx}
\usepackage{subfigure}
\usepackage{booktabs} 

\usepackage{hyperref}

\usepackage[accepted]{icml2023}

\usepackage{amsmath}
\usepackage{amssymb}
\usepackage{mathtools}
\usepackage{amsthm}

\usepackage{url}

\usepackage{multirow}

\usepackage{makecell}
\usepackage[group-separator={,}]{siunitx}
\usepackage{xfrac}

\usepackage[capitalize,noabbrev]{cleveref}

\theoremstyle{plain}
\newtheorem{theorem}{Theorem}[section]

\newtheorem{lemma}[theorem]{Lemma}

\theoremstyle{definition}
\newtheorem{definition}[theorem]{Definition}
\newtheorem{assumption}[theorem]{Assumption}
\theoremstyle{remark}
\newtheorem{remark}[theorem]{Remark}

\usepackage[textsize=tiny]{todonotes}

\usepackage{amsmath,amsfonts,bm}

\def\eqref#1{equation~\ref{#1}}

\def\1{\bm{1}}

\def\rvz{{\mathbf{z}}}

\def\va{{\bm{a}}}
\def\vb{{\bm{b}}}

\def\vp{{\bm{p}}}
\def\vq{{\bm{q}}}
\def\vr{{\bm{r}}}

\def\vw{{\bm{w}}}
\def\vx{{\bm{x}}}
\def\vy{{\bm{y}}}
\def\vz{{\bm{z}}}

\def\mH{{\bm{H}}}

\def\mJ{{\bm{J}}}

\def\mW{{\bm{W}}}

\DeclareMathAlphabet{\mathsfit}{\encodingdefault}{\sfdefault}{m}{sl}
\SetMathAlphabet{\mathsfit}{bold}{\encodingdefault}{\sfdefault}{bx}{n}

\def\sD{{\mathbb{D}}}

\def\sM{{\mathbb{M}}}

\def\sS{{\mathbb{S}}}

\icmltitlerunning{FADE: Enabling Federated Adversarial Training on Heterogeneous Resource-Constrained Edge Devices}

\begin{document}

\twocolumn[
\icmltitle{FADE: Enabling Federated Adversarial Training \\on Heterogeneous Resource-Constrained Edge Devices}



\icmlsetsymbol{equal}{*}

\begin{icmlauthorlist}
\icmlauthor{Minxue Tang}{duke}
\icmlauthor{Jianyi Zhang}{duke}
\icmlauthor{Louis DiValentin}{acc}
\icmlauthor{Aolin Ding}{acc}
\icmlauthor{Amin Hassanzadeh}{acc}
\icmlauthor{Hai Li}{duke}
\icmlauthor{Yiran Chen}{duke}
\end{icmlauthorlist}

\icmlaffiliation{duke}{Department of ECE, Duke University}
\icmlaffiliation{acc}{Accenture}

\icmlcorrespondingauthor{Yiran Chen}{yiran.chen@duke.edu}

\icmlkeywords{Machine Learning, ICML}

\vskip 0.3in
]



\printAffiliationsAndNotice{}  

\begin{abstract}
Federated adversarial training can effectively complement adversarial robustness into the privacy-preserving federated learning systems. However, the high demand for memory capacity and computing power makes large-scale federated adversarial training infeasible on resource-constrained edge devices. Few previous studies in federated adversarial training have tried to tackle both memory and computational constraints simultaneously. In this paper, we propose a new framework named Federated Adversarial Decoupled Learning (FADE) to enable AT on heterogeneous resource-constrained edge devices. FADE differentially decouples the entire model into small modules to fit into the resource budget of each device, and each device only needs to perform AT on a single module in each communication round. We also propose an auxiliary weight decay to alleviate objective inconsistency and achieve better accuracy-robustness balance in FADE. FADE offers theoretical guarantees for convergence and adversarial robustness, and our experimental results show that FADE can significantly reduce the consumption of memory and computing power while maintaining accuracy and robustness.
\end{abstract}
\section{Introduction}\label{Sec:Intro}
As a privacy-preserving distributed learning paradigm, Federated Learning (FL) makes a meaningful step toward the practice of secure and trustworthy artificial intelligence~\cite{konevcny2015federated,konevcny2016federated,mcmahan2017communication,kairouz2019advances}. In contrast to traditional centralized training, FL pushes the training to edge devices (clients), and client models are locally trained and uploaded to the server for aggregation. Since no private data is shared with other clients or the server, FL substantially improves data privacy during the training process.

While FL can preserve the privacy of the participants, other threats can still impact the reliability of the machine learning model running on the FL system. 
One such threat is adversarial examples that aim to cause misclassifications by adding imperceptible noise into the input data~\cite{szegedy2013intriguing,goodfellow2014explaining}. Previous research has shown that performing adversarial training (AT) on a large model is an effective method to attain robustness against adversarial examples while maintaining high accuracy on clean data~\cite{liu2020adversarial}.
However, large-scale AT also puts high demand for both memory capacity and computing power. Thus it becomes unaffordable for some edge devices with limited resources, such as mobile phones and IoT devices that contribute to the majority of the participants in cross-device FL~\cite{kairouz2019advances,li2020federated,wong2020fast,zizzo2020fat,hong2021federated}. \cref{tab:part_at} shows that strong robustness of the whole FL system cannot be attained by allowing only a small portion (e.g., $20\%$) of the clients to perform AT. Therefore, enabling resource-constrained edge devices to perform AT is necessary for achieving strong robustness in FL.

\begin{table*}[t]
\centering
\caption{Results of partial federated adversarial training with 100 clients. ``$20\%$ AT + $80\%$ ST'' means that $20\%$ clients perform AT while $80\%$ clients perform standard training (ST).}
\label{tab:part_at}
\begin{tabular}{ccc|cc}
\hline
\multirow{2}{*}{Training Scheme} & \multicolumn{2}{c|}{FMNIST (CNN-7)}     & \multicolumn{2}{c}{CIFAR-10 (VGG-11)}    \\ \cline{2-5} 
                                 & Natural Acc. & Adversarial Acc. & Natural Acc. & Adversarial Acc. \\ \hline
$100\%$ AT + $0\%$ ST            & $78.39\%$    & $66.93\%$        & $64.73\%$        & $33.27\%$            \\
$20\%$ AT + $80\%$ ST            & $83.83\%$    & $48.61\%$        & $74.77\%$        & $19.22\%$            \\ \hline
\end{tabular}
\vspace{-1em}
\end{table*}
Some previous works have tried to tackle client-wise systematic heterogeneity in FL~\cite{li2018federated,xie2019asynchronous,lu2020privacy,wang2020tackling}. The most common method is allowing slow devices to perform fewer epochs of local training than the others~\cite{li2018federated,wang2020tackling}. While this method can reduce the computational costs on resource-constrained devices, the memory capacity limitation has not been well addressed in these works.

To tackle both memory capacity and computing power constraints, recent studies propose a novel training scheme named Decoupled Greedy Learning (DGL) which partitions the entire neural network into multiple small modules and trains each module separately~\cite{belilovsky2019greedy,wang2021revisiting}. Although DGL successfully reduces both memory and computational requirements for training large models, the exploration and convergence analysis of DGL are limited to the case that all computing nodes adopt the same partition of the model~\cite{belilovsky2020decoupled}, which cannot fit into different resource budgets of different clients in heterogeneous FL. Additionally, no previous studies have discussed whether DGL can be combined with AT to confer joint adversarial robustness to the entire model. It is not trivial to achieve joint robustness of the entire model when applying AT with DGL, since DGL trains each module separately with different locally supervised losses.

In this paper, we propose \textbf{F}ederated \textbf{A}dversarial \textbf{DE}coupled Learning (FADE), which is the first adversarial decoupled learning framework for heterogeneous FL with convergence and robustness guarantees. 
Our main contributions are:

\begin{enumerate}
    \item We propose Federated Decoupled Learning (FDL) allowing differentiated model partitions on heterogeneous devices with different resource budgets. We give a theoretical guarantee for the convergence of FDL.
    \item We propose \textbf{F}ederated \textbf{A}dversarial \textbf{DE}coupled Learning (FADE) to attain theoretically guaranteed joint adversarial robustness of the entire model. Our experimental results show that FADE can significantly reduce the memory and computational requirements while maintaining almost the same natural accuracy and adversarial robustness as joint training.
    \item We reveal the non-trivial relationship between objective consistency (natural accuracy) and adversarial robustness in FADE, and we propose an effective method to achieve a better accuracy-robustness balance point with the weight decay on auxiliary models.
\end{enumerate}

\section{Preliminary}\label{Sec:preliminary}
\paragraph{Federated Learning (FL)} In FL, different clients collaboratively train a shared global model $\vw$ with locally stored data~\cite{mcmahan2017communication}. The objective of FL is:
\begin{align}
\min_{\vw}&\quad L(\vw)=\sum_{k=1}^N q_kL_k(\vw),\\ 
\text{where}&\quad L_k(\vw) = \frac{1}{\vert\sD_k\vert}\sum_{(\vx,y)\in \sD_k} l(\vx,y;\vw),
\end{align}
and $l$ is the task loss, e.g., cross-entropy loss for classification tasks. $\sD_k$ is the dataset of client $k$ and its weight $q_k=\vert \sD_k\vert/(\sum_i \vert \sD_i\vert)$. To solve for the optimal solution of this objective, in each communication round, FL first samples a subset of clients $\sS^{(t)}$ to perform local training. These clients initialize their models with the global model $\vw_k^{(t,0)}=\vw^{(t)}$, and then run $\tau$ iterations of local SGD. After all these clients complete training in this round, their models are uploaded and averaged to become the new global model~\cite{mcmahan2017communication}. We summarize this procedure as follows:
\begin{align}
    \vw_k^{(t+1)}&=\vw^{(t)} - \eta_t\sum_{j=0}^{\tau-1}\nabla L_k(\vw_k^{(t,j)}),\\
    \vw^{(t+1)}&=\frac{1}{\sum_{i\in\sS^{(t)}} q_i}\sum_{k\in\sS^{(t)}} q_k \vw_k^{(t+1)},
\end{align}
where $\vw_k^{(t,j)}$ is the local model of client $k$ at the $j$-th iteration of communication round $t$. 

\paragraph{Adversarial Training (AT)} The goal of AT is to achieve robustness against small perturbations in the inputs. We define $(\epsilon,c)$-robustness as follows: 
\begin{definition}
We say a model $\vw$ is $(\epsilon,c)$-robust in a loss function $l$ at input $\vx$ if $\forall \boldsymbol{\delta}\in\{\boldsymbol{\delta}:\Vert \boldsymbol{\delta} \Vert_p \le \epsilon\}$,
\begin{align}
    l(\vx+\boldsymbol{\delta},y;\vw)-l(\vx,y;\vw) \le c,
\end{align}
where $\Vert\cdot\Vert_p$ is the $\ell_p$ norm of a vector\footnote{For simplicity, without specifying $p$, we use $\Vert\cdot\Vert$ for $\ell_2$ norm. Our conclusions in the following sections can be extended to any $\ell_p$ norm with the equivalence of vector norms.}, and $\epsilon$ is the perturbation tolerance.
\end{definition}
AT trains a model with adversarial examples to achieve adversarial robustness, which can be formulated as a min-max problem~\cite{goodfellow2014explaining,madry2017towards}:
\begin{align}
    \min_{\vw}\max_{\boldsymbol{\delta}:\Vert \boldsymbol{\delta} \Vert_p\le \epsilon} l(\vx+\boldsymbol{\delta},y;\vw).\label{eq:at}
\end{align}
To solve \cref{eq:at}, people usually alternatively solve the inner maximization and the outer minimization. While solving the inner maximization, Projected Gradient Descent (PGD) is shown to introduce the strongest robustness in AT \cite{madry2017towards,wong2020fast,wang2021revisiting}. 

\paragraph{Decoupled Greedy Learning (DGL)} The key idea of DGL is to partition the entire model into multiple non-overlapping small modules. By introducing a locally supervised loss to each module, we can load and train each module independently without accessing the other parts of the entire model~\cite{belilovsky2019greedy,belilovsky2020decoupled}. This enables devices with small memory to train large models.

\begin{figure}[t]
\centering
\includegraphics[width=1.0\linewidth]{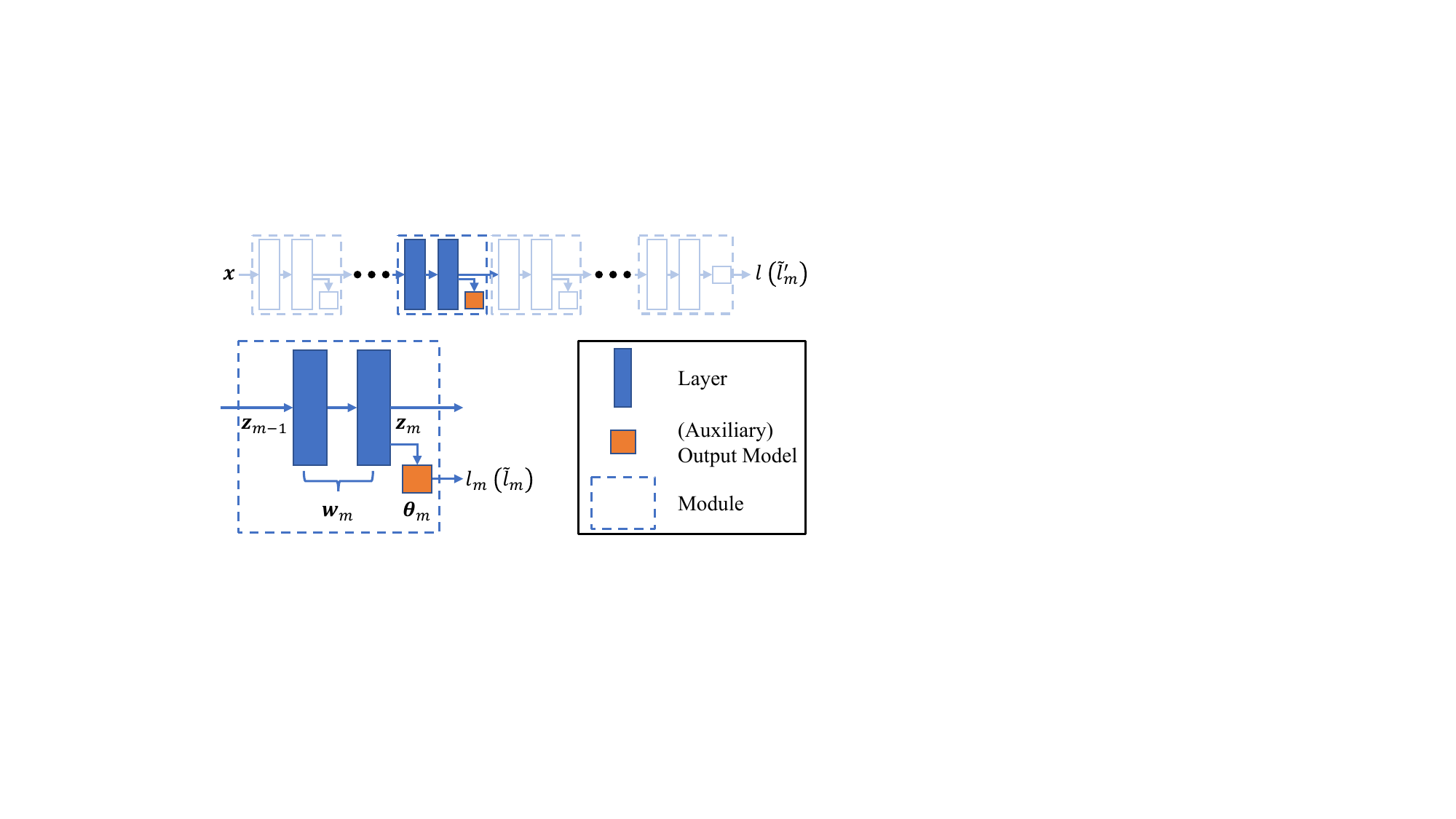}
\vspace{-1.5em}
\caption{An illustration of DGL and a single module $m$.}\label{fig:module}
\vspace{-0.5em}
\end{figure}

As shown in \cref{fig:module}, each module $m$ contains one or multiple adjacent layers $\vw_m$ of the backbone neural network, together with a small auxiliary model $\boldsymbol{\theta}_m$ that provides locally supervised loss. We denote $\boldsymbol{\Theta}_m=[\vw_m,\boldsymbol{\theta}_m]$ to be all the parameters in module $m$. Module $m$ accepts the features $\vz_{m-1}$ from the previous module as the input, and it outputs features $\vz_{m}=f_m(\vz_{m-1};\vw_{m})$ for the following modules, as well as a locally supervised loss $l_m(\vz_{m-1},y;\boldsymbol{\Theta}_m)$. At epoch $t$, the averaged locally supervised loss $L_m^{(t)}$ will be used for training this module:
\begin{align}
    L_m^{(t)}(\boldsymbol{\Theta}_m^{(t)})=\mathbb{E}_{(\vz_{m-1}^{(t)},y)} \left[l_m(\vz_{m-1}^{(t)},y;\boldsymbol{\Theta}_m^{(t)})\right].\label{eq:eeloss}
\end{align}
In contrast to joint training, the input of one module can be various in different epochs in DGL since we may keep updating the previous modules during training. Thus we use $\vz_{m-1}^{(t)}$ to denote the inputs of module $m$ in epoch $t$, and only the input of the first module $\vz_0^{(t)}=\vx$ is invariant.

For convenience, we define the loss function of the auxiliary model $\boldsymbol{\theta}_m$ and the loss function of the following layers $[\vw_{m+1},\cdots,\vw_{M}]$ in the backbone network as 
\begin{align}
    &\tilde l_m(\vz_m,y;\boldsymbol{\theta}_m)=l_m(\vz_{m-1},y;\boldsymbol{\Theta}_m),\\
    &\tilde l'_m(\vz_m,y;\vw_{m+1},\cdots,\vw_{M})=l(\vx,y;\vw),
\end{align}
for each module $m$. Without specifying, we will omit all parameters ($\vw_m,\boldsymbol{\theta}_m$ and $\boldsymbol{\Theta}_m$) in the following sections.

\section{Federated Adversarial Decoupled Learning}\label{Sec:method}
In this section, we present Federated Adversarial Decoupled Learning (FADE), which aims at enabling all clients with different computing resources to participate in adversarial training. In \cref{Subsec:fdl}, we introduce Federated Decoupled Learning (FDL) with differentiated model partitions for heterogeneous resource-constrained devices, and we analyze its convergence property. In \cref{Subsec:advdgl}, we integrate AT into FDL to achieve guaranteed joint adversarial robustness of the entire model. In \cref{Subsec:wd}, we discuss the objective inconsistency in FDL and propose an effective method to attain a better accuracy-robustness balance point.

\subsection{Federated Decoupled Learning}\label{Subsec:fdl}
\begin{figure*}[t]
\centering
\includegraphics[width=0.9\linewidth]{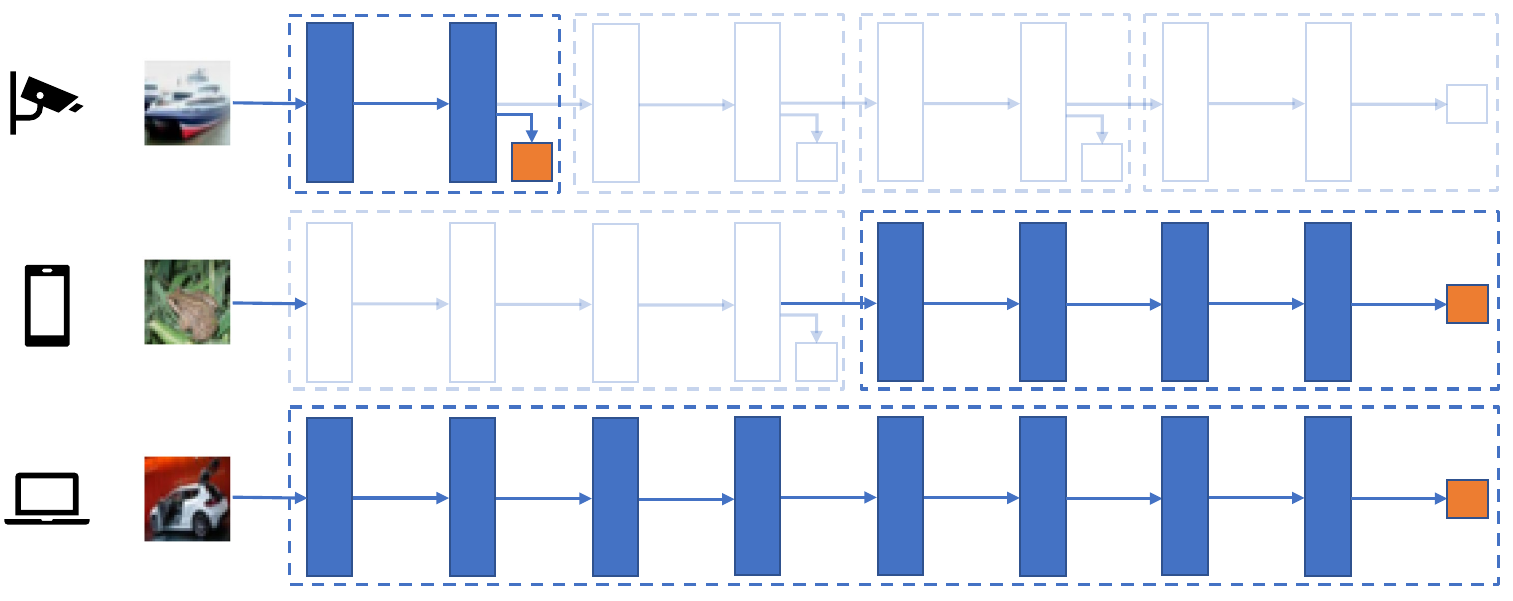}
\vspace{-1.2em}
\caption{A framework of Federated Decoupled Learning. In contrast to using one unified partition in DGL~\cite{belilovsky2020decoupled}, we allow different model partitions among devices according to their resource budgets. In each communication round, each device randomly selects one module (highlighted) for training, and then the updates of each layer will be averaged respectively (column-wise in the figure).}
\label{fig:fdl}
\vspace{-0.6em}
\end{figure*}

In cross-device FL, the main participants are usually small edge devices that have limited hardware resources and may not be able to afford large-scale AT that requires large memory and high computing power~\cite{li2018federated,kairouz2019advances,li2020federated,wang2020tackling}. A solution to tackle the resource constraints on edge devices is to deploy DGL in FL, such that each device only needs to load and train a single small module instead of the entire model in each communication round. However, DGL only allows a unified model partition on all the devices~\cite{belilovsky2020decoupled}. Considering the systematic heterogeneity, we would prefer differentiated model partitions to fit into different resource budgets of different clients. As shown in \cref{fig:fdl}, devices with limited resources (such as IoT devices) can partition the entire model into smaller modules, and devices with more resources (such as mobile phones or computers) can train larger modules or even the entire model. 

Accordingly, we propose Federated Decoupled Learning (FDL) to tackle heterogeneous resource budgets. We define the model partition $\sM_k$ as the set of all modules on client $k$. In contrast to DGL, since various model partitions $\sM_k$ can be used by different clients in FDL, a module on one client may not be a module on the other clients. Thus instead of a specific module, we consider the aggregation rule of each layer $n$ with parameter $\boldsymbol{\omega}_n$ (either in the backbone network or in the auxiliary models), as one single layer is the ``atom'' in FDL and cannot be further partitioned. We use $m_k(n)$ to denote the module that contains layer $n$ on client $k$, and we define $L_{n,k}=L_{m_k(n),k}$ as the locally supervised loss for training this layer. In each communication round $t$, each client $k$ randomly samples a module $m_k^t$ from $\sM_k$ for training (\cref{eq:fdl}). After the local training, the server averages the updates of each layer $n$ over clients whose trained modules contain layer $n$ in this round, i.e., clients in $\sS_{n}^{(t)}=\{k\in\sS^{(t)}:n \in m_k^t\}$ (\cref{eq:aggre}).
\begin{align}
&\boldsymbol{\omega}_{n,k}^{(t+1)}=
\begin{cases}
\boldsymbol{\omega}_n^{(t)}-\eta_t\sum_{j}\nabla_{\boldsymbol{\omega}_n} L_{n,k}^{(t)}
,\ \text{ if }n \in m_k^t;\\
\boldsymbol{\omega}_n^{(t)},\qquad\qquad\qquad\qquad\ \text{elsewhere}.
\end{cases}\label{eq:fdl}\\
&\boldsymbol{\omega}_n^{(t+1)}=\frac{1}{\sum_{i\in \sS_n^{(t)}}q_i}\sum_{k\in\sS_n^{(t)}}q_k\boldsymbol{\omega}_{n,k}^{(t+1)}.\label{eq:aggre}
\end{align}

\cref{theorem:convergence} guarantees the convergence of FDL, while the full version with proof is given in \cref{Apx:convergence}.
\begin{theorem}~\label{theorem:convergence}
Under some common assumptions, the locally supervised loss $L_n=\sum_k q_kL_{n,k}$ of any layer $n$ can converge in Federated Decoupled Learning: 
\begin{align}
\lim_{T\rightarrow \infty}\inf_{t \leq T} \mathbb{E}\left\|\nabla_{\boldsymbol{\omega}_n} L_n\right\|^{2} = 0.
\end{align}
\end{theorem}

\cref{theorem:convergence} can guarantee the convergence of locally supervised loss $L_n$. However, because of the objective inconsistency $\Vert \nabla L-\nabla L_n\Vert\ge 0$, we cannot guarantee the convergence of the joint loss $L$ with this result. We will discuss the objective inconsistency in \cref{Subsec:wd}, and we will show how we can reduce this gap such that we can make the joint loss gradient $\nabla L$ smaller when the locally supervised loss $L_n$ converges.

\subsection{Adversarial Decoupled Learning}\label{Subsec:advdgl}
Now we discuss how to integrate AT into FDL for joint robustness of the entire model. We consider the training on a single client in this section and the next, thus we omit client $k$ and use module $m$ instead of layer $n$ in the subscripts.

We can obtain adversarial decoupled learning by replacing the standard locally supervised loss of each module $m$ with the adversarial loss as follows:
\begin{align}
    \min_{\boldsymbol{\Theta}_m}\max_{\Vert\boldsymbol{\delta}_{m-1}\Vert\le \epsilon_{m-1}}\quad l_m(\vz_{m-1}+\boldsymbol{\delta}_{m-1},y;\boldsymbol{\Theta}_m).
\end{align}
However, two concerns have not been addressed in adversarial decoupled learning:
\begin{enumerate}
    \item Since different modules are trained with different locally supervised losses, can the local robustness in $l_m$ of each module guarantee the joint robustness in $l$ of the entire (backbone) model?
    \item When applying AT on a module $m$, what value of the perturbation tolerance $\epsilon_{m-1}$ should we use to ensure the joint robustness of the entire model?
\end{enumerate}

\cref{theorem:robustness} reveals the relationship between the local robustness of each module and the joint robustness of the entire model, and it gives a lower bound of the perturbation tolerance $\epsilon_{m-1}$ for each module $m$ to sufficiently guarantee the joint robustness. \cref{theorem:robustness} is proved in \cref{pf:theorem:robustness}.

\begin{theorem}\label{theorem:robustness}
Assume that $\tilde l_m(\vz_m,y)$ is $\mu_m$-strongly convex in $\vz_m$ for each module $m$. We can guarantee that the entire model has a joint $(\epsilon_0,c_M)$-robustness in $l(\vx,y)$, if each module $m\le M$ has local $(\epsilon_{m-1},c_m)$-robustness in $l_m(\vz_{m-1},y)$, and
\begin{align}
    \quad\epsilon_{m}\ge \frac{g_m}{\mu_m}+\sqrt{\frac{2c_m}{\mu_m}+\frac{g_m^2}{\mu_m^2}},\label{eq:eps_lb}
\end{align}
where $g_m=\left\|\nabla_{\vz_m}\tilde l_m(\vz_m,y)\right\|$.
\end{theorem}

\begin{remark} 
In \cref{theorem:robustness}, we assume that the loss function $\tilde l_m(\vz_m,y)$ of the auxiliary model $\boldsymbol{\theta}_m$ is strongly convex in its input $\vz_m$. This assumption is realistic since the auxiliary model $\boldsymbol{\theta}_m$ is usually a very simple model, e.g., only a linear layer followed by cross-entropy loss. We also theoretically analyze the sufficiency of a simple auxiliary model in \cref{Subsec:wd} (See \cref{remark2}).
\end{remark}

\cref{theorem:robustness} shows that larger $\mu_m$ and smaller $g_m$ will lead to stronger joint robustness of the entire model since the lower bound of $\epsilon_m$ becomes smaller for ensuring the joint robustness. Both $\mu_m$ and $g_m$ are related to $\tilde l_m$ parameterized by the auxiliary model $\boldsymbol{\theta}_m$. We will discuss how we can regularize the auxiliary model to attain better robustness while reducing the objective inconsistency of FDL in \cref{Subsec:wd}.

\subsection{Auxiliary Weight Decay}\label{Subsec:wd}
\begin{figure}[t]
\centering
\includegraphics[width=0.9\linewidth]{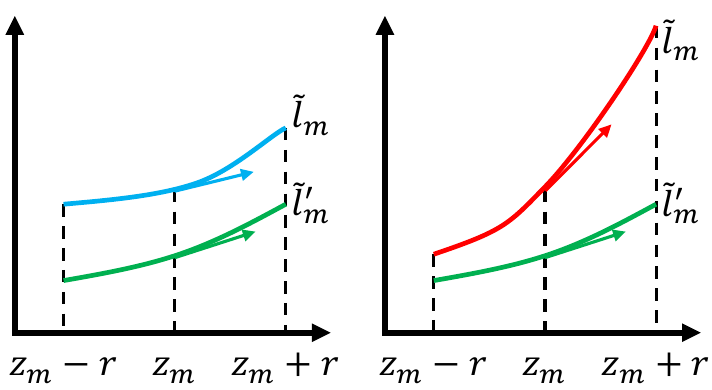}
\vspace{-0.5em}
\caption{An illustration of the relationship between objective consistency and robustness. If both $\tilde l_m$ and $\tilde l_m'$ are robust and smooth as shown in the left figure, their gradients are both close to $\boldsymbol{0}$ and the difference between the gradients is small. If $\tilde l_m$ is not robust, the difference between gradients is large in the right figure.}\label{fig:illust}
\vspace{-0.5em}
\end{figure}

As we mentioned in \cref{Subsec:fdl}, there exists objective inconsistency in FDL because each module is trained with the locally supervised loss instead of the joint loss. The objective inconsistency is defined as $\Vert \nabla_{\vw_m} l-\nabla_{\vw_m} l_m\Vert$, which is the difference between the gradients of the joint loss and the locally supervised loss. The existence of this inconsistency makes the optimal parameters that minimize the locally supervised loss $l_m$ not necessarily minimize the joint loss $l$. Furthermore, the objective inconsistency can enlarge heterogeneity among clients and hinder the convergence of FL~\cite{li2019convergence,wang2020tackling}. Therefore, it is important to alleviate the objective inconsistency in FDL to improve its convergence and performance.

\cref{theorem:inconsistency} shows a non-trivial relationship between objective inconsistency and adversarial robustness: strong joint adversarial robustness also implies small objective inconsistency in FDL. We prove \cref{theorem:inconsistency} in \cref{pf:theorem:inconsistency}.
\begin{lemma}\label{theorem:inconsistency}
Assume that $\tilde l_m(\vz_m,y)$ and $\tilde l'_m(\vz_{m},y)$ are $\beta_m,\beta'_{m}$-smooth in $\vz_m$ for a module $m$. If there exist $\tilde c_m$, $\tilde c'_{m}$, and $r\ge \sqrt{2\frac{\tilde c_m+\tilde c'_{m}}{\beta_m+\beta'_{m}}}$, such that the auxiliary model has $(r,\tilde c_m)$-robustness in $\tilde l_m(\vz_m,y)$, and the backbone network has $(r,\tilde c'_m)$-robustness in $\tilde l'_m(\vz_m,y)$, then we have:
\begin{align}
    \Vert\nabla_{\vw_m} l-&\nabla_{\vw_m} l_m\Vert \le \left\Vert \frac{\partial \vz_m}{\partial \vw_m}\right\Vert\Vert \nabla_{\vz_m} \tilde l_m'-\nabla_{\vz_m} \tilde l_m\Vert\nonumber\\
    \le &\left\Vert \frac{\partial \vz_m}{\partial \vw_m}\right\Vert\sqrt{2(\tilde c_m+\tilde c'_{m})(\beta_m+\beta'_{m})}.\label{eq:theorem:inconsistency}
\end{align}
\end{lemma}

\cref{theorem:inconsistency} suggests that we can alleviate the objective inconsistency by reducing $\beta_m,\beta'_m,\tilde c_m$ and $\tilde c_m'$ (Regularizing $\Vert \partial \vz_m/\partial \vw_m\Vert$ usually requires second derivative, which introduces high memory and computational overhead, so we do not consider it here). Notice that adversarial decoupled learning can guarantee an ($\epsilon_{m},c_M$)-robustness in $\tilde l'_m$ according to \cref{theorem:robustness}, which implies a small $\tilde c_m'$. Furthermore, \citet{moosavi2019robustness} shows that adversarial robustness also implies a smoother loss function. Therefore, adversarial decoupled learning also leads to a small $\beta'_m$. 

Accordingly, with adversarial decoupled learning, reducing $\beta_m$ and $\tilde c_m$ can effectively alleviate the objective inconsistency, which is also illustrated in \cref{fig:illust}. We notice that both $\beta_m$ and $\tilde c_m$ are only related to $\tilde l_m$ parameterized by the auxiliary model $\boldsymbol{\theta}_m$, and \cref{Apx:case} shows that we can reduce $\beta_m$ and $\tilde c_m$ by adding a large weight decay on the auxiliary model when it is simple (e.g., only a linear layer).

\begin{remark} \label{remark2}
It is noteworthy that we do not use any conditions on the difference between $\tilde l'_m$ and $\tilde l_m$ in both \cref{theorem:robustness} and \cref{theorem:inconsistency} (also \cref{fig:illust}). This implies that the auxiliary model is not required to perform as well as the joint backbone model. Thus, a simple auxiliary model is sufficient to achieve high joint robustness and low objective inconsistency in adversarial decoupled learning.
\end{remark}

Based on all analysis in this work, we propose Federated Adversarial Decoupled Learning (FADE), where we replace the original loss function $l_m$ used by FDL with the following adversarial loss with weight decay on the auxiliary model:
\begin{equation}
\begin{aligned}
    &l_m^{\text{FADE}}(\vz_{m-1}^{(t)},y;\vw_{m}^{(t)},\boldsymbol{\theta}_m^{(t)})\\
    = &\max_{\boldsymbol{\delta}_{m-1}^{(t)}}\left[ l_m(\vz_{m-1}^{(t)}+\boldsymbol{\delta}_{m-1}^{(t)},y;\vw_{m}^{(t)},\boldsymbol{\theta}_m^{(t)})\right]+\lambda_m\Vert \boldsymbol{\theta}_m^{(t)}\Vert^2,\label{eq:fade_loss}
\end{aligned}
\end{equation}
where $\lambda_m$ is the hyperparameter that controls the weight decay on the auxiliary model $\boldsymbol{\theta}_m$. Our framework is summarized in \cref{alg:fade}.

\begin{algorithm}[tb]
\caption{Federated Adversarial Decoupled Learning}
\label{alg:fade}
\begin{algorithmic}[1]
\STATE Initialize $\vw^{(0)}$ and $\boldsymbol{\theta}_m^{(0)}$ for each module $m$.
\FOR{$t=1,2,\cdots, T$}
    \STATE Randomly sample a group of clients $\sS^{(t)}$ for training.
    \FOR{each client $k\in\sS^{(t)}$ in parallel}
        \STATE Randomly selects a module $m_k^t$.
        \STATE Download $\boldsymbol{\Theta}_{m_k^t}^{(t)}=[\vw_{m_k^t}^{(t)},\boldsymbol{\theta}_{m_k^t}^{(t)}]$ from the server.
        \STATE Generate input features $\vz_{m_k^t-1}^{(t)}$ for module $m_k^t$. 
        \STATE Perform AT on module $m_k^t$ with $l_{m_k^t}^{\text{FADE}}$ in \cref{eq:fade_loss} for $\tau$ iterations, and get $\boldsymbol{\Theta}_{m_k^t,k}^{(t+1)}$.
        \STATE Upload $\boldsymbol{\Theta}_{m_k^t,k}^{(t+1)}$ to the server.
    \ENDFOR
\STATE The server aggregates $\boldsymbol{\omega}_{n,k}^{(t+1)}$ to get $\boldsymbol{\omega}_n^{(t+1)}$ according to \cref{eq:aggre} for each layer $n$.
\ENDFOR
\end{algorithmic}
\end{algorithm}

\begin{figure*}[t]
\centering
\includegraphics[width=1.0\linewidth]{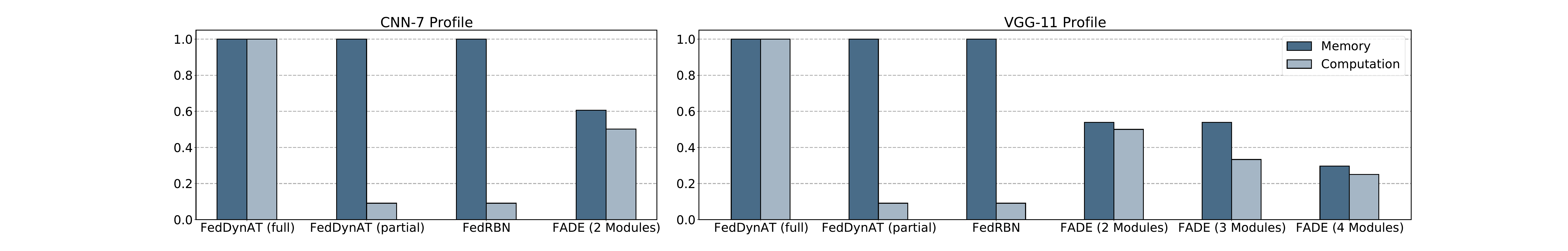}
\vspace{-2em}
\caption{Minimum memory and computational requirements of baselines and FADE with different numbers of modules. The results are shown as the percentage of the resource requirement of full FedDynAT with PGD-10 AT.}
\label{fig:profile}
\vspace{-0.5em}
\end{figure*}

\paragraph{Trade-off Between Joint Accuracy and Joint Robustness.}As we discussed in \cref{Subsec:advdgl} and this section, four parameters ($\mu_m,g_m,\tilde c_m$ and $\beta_m$) that are only related to the auxiliary model $\boldsymbol{\theta}_m$ can influence the joint robustness and objective consistency. We can see in \cref{Apx:case} that applying a larger $\lambda_m$ can decrease all of them. Smaller $\tilde c_m$ and $\beta_m$ can alleviate the objective inconsistency to increase the joint accuracy, and smaller $g_m$ can improve the joint robustness. However, smaller $\mu_m$ will lead to weaker robustness by increasing the lower bound of $\epsilon_m$. Therefore, there exists an accuracy-robustness trade-off when we apply the weight decay, and the value of $\lambda_m$ plays an important role in balancing the joint accuracy and the joint robustness.

\section{Experimental Results}\label{Sec:exp}
\subsection{Experiment Settings}

We conduct our experiments on two datasets, FMNIST~\cite{xiao2017fashion} and CIFAR-10 \cite{krizhevsky2009learning}, partitioned onto $N=100$ clients with the same Non-IID data partition as \citet{shah2021adversarial}. We sample $30$ clients for local training in each communication round.

We conduct two groups of experiments with two different FL optimizers respectively: FedNOVA~\cite{wang2020tackling} for global FL and FedBN~\cite{li2021fedbn} for personalized FL. Notice that the results in global FL and personalized FL are not comparable since they assume different test set partitions. We combine FADE with different FL optimizers to show the generalization of our method.

For FMNIST, we use a $7$-layer CNN (CNN-7) with five convolutional layers and two fully connected layers. We adopt two model partitions with 1 or 2 modules for CNN-7 in FADE. For CIFAR-10, we use VGG-11~\cite{simonyan2014very} as the model. We adopt four different model partitions for VGG-11 in FADE, with 1, 2, 3 or 4 modules respectively. We use a linear layer as the auxiliary model for each module in FADE.


For AT settings, we use $l_\infty$ norm to bound the perturbation and use PGD-10 to generate adversarial examples for training and testing, following \citet{zizzo2020fat}.


we will compare FADE with three baselines. Full FedDynAT (``FedDynAT ($100\%$ AT)'')~\cite{shah2021adversarial} represents the ideal performance of federated adversarial training when all the clients are able to perform joint AT on the entire model. While full FedDynAT is not feasible under our constraint that only a small portion of clients can afford joint AT, we adopt partial FedDynAT as the baseline where clients with insufficient resources only perform joint standard training (ST). Another baseline FedRBN~\cite{hong2021federated} also allows resource-constrained devices performing joint ST only, and the robustness will be propagated by transferring the batch-normalization statistics from the clients who can afford joint AT to the clients who only perform joint ST. 

We provide detailed experiment settings and introductions to the baselines in \cref{Apx:exp_detail}.


\subsection{Resource Requirements}
\begin{table*}[t]
\centering
\caption{The natural accuracy (clean examples) and adversarial accuracy (adversarial examples) on FMNIST. Results are reported in the mean and the standard deviation over three random seeds. See \cref{table:hyperparameters} for $\epsilon_m$ and $\lambda_m$ settings of FADE.}
\label{table:fade_fmnist_res}
\begin{tabular}{ccc|cc}
\hline
\multirow{2}{*}{Training Scheme}        & \multicolumn{2}{c|}{FedNOVA}     & \multicolumn{2}{c}{FedBN}    \\ \cline{2-5} 
                                        & Natural Acc. & Adversarial Acc. & Natural Acc. & Adversarial Acc. \\ \hline
FedDynAT ($100\%$ AT)                   & $78.39\pm0.65\%$    & $66.93\pm0.87\%$        & $89.85\pm0.41\%$    & $82.92\pm0.66\%$        \\ \hline
FedDynAT ($20\%$ AT)                    & $83.83\pm0.32\%$    & $48.61\pm0.87\%$        & $91.94\pm0.07\%$    & $16.02\pm1.03\%$        \\ 
FedRBN                                  & n/a                 & n/a                     & $90.35\pm1.51\%$    & $62.14\pm6.45\%$        \\ \hline
FADE (2:8)                        & $78.74\pm1.09\%$    & $66.72\pm2.09\%$        & $89.43\pm0.52\%$    & $81.24\pm0.94\%$        \\ \hline
\end{tabular}
\vspace{-1em}
\end{table*}

\begin{table*}[t]
\centering
\caption{The natural accuracy (clean examples) and adversarial accuracy (adversarial examples) on CIFAR-10. Results are reported in the mean and the standard deviation over three random seeds. See \cref{table:hyperparameters} for $\epsilon_m$ and $\lambda_m$ settings of FADE.}
\label{table:fade_cifar_res}
\begin{tabular}{ccc|cc}
\hline
\multirow{2}{*}{Training Scheme}        & \multicolumn{2}{c|}{FedNOVA}     & \multicolumn{2}{c}{FedBN}    \\ \cline{2-5} 
                                        & Natural Acc. & Adversarial Acc. & Natural Acc. & Adversarial Acc. \\ \hline
FedDynAT ($100\%$ AT)                   & $64.73\pm1.63\%$    & $33.27\pm0.46\%$        & $81.71\pm0.14\%$    & $57.28\pm1.23\%$        \\ \hline
FedDynAT ($20\%$ AT)                    & $74.77\pm1.68\%$    & $19.22\pm2.16\%$        & $87.12\pm0.25\%$    & $16.51\pm1.64\%$        \\ 
FedRBN                                  & n/a                 & n/a                     & $86.80\pm0.31\%$    & $53.08\pm1.03\%$        \\ \hline
FADE (2:8:0:0)                        & $65.42\pm0.42\%$    & $32.22\pm0.43\%$        & $81.05\pm0.56\%$    & $59.12\pm0.63\%$        \\
FADE (2:0:8:0)                        & $64.72\pm0.68\%$    & $31.81\pm0.35\%$        & $77.46\pm0.67\%$    & $58.14\pm0.85\%$        \\ 
FADE (2:0:0:8)                        & $63.19\pm0.72\%$    & $29.33\pm0.20\%$        & $78.72\pm0.50\%$    & $57.37\pm0.50\%$        \\ 
FADE (2:3:5:0)                          & $66.06\pm1.09\%$    & $32.28\pm0.49\%$        & $78.23\pm0.35\%$    & $58.80\pm0.66\%$        \\
FADE (2:3:0:5)                          & $64.79\pm0.48\%$    & $30.98\pm0.52\%$        & $79.07\pm0.40\%$    & $57.28\pm0.96\%$        \\
FADE (2:2:3:3)                          & $65.53\pm1.10\%$    & $32.03\pm0.37\%$        & $77.80\pm0.39\%$    & $57.47\pm0.42\%$        \\\hline
\end{tabular}
\end{table*}
\cref{fig:profile} shows the minimum resource requirements of the baselines and FADE with multi-module model partitions. We use the number of loaded parameters as the metric of memory, and the FLOPs as the metric of computation. For partial FedDynAT and FedRBN, the minimum memory requirement is the number of parameters in the entire model since they always load the entire model for training, and the computing power requirement is the FLOPs of ST on the entire model since the resource-constrained devices only perform ST. For FADE, the minimum memory requirement is the number of parameters in the largest module, while the computing power requirement is the mean of FLOPs for PGD-10 AT in each module.

Among all the different methods, we can see that only FADE can reduce the memory requirement for training. FADE with 2 modules can reduce the memory requirement by more than $40\%$ on both CNN-7 and VGG-11, while FADE with 4 modules can further reduce the memory requirement by more than $70\%$. At the same time, FADE reduces the computation by $50\%$ to $75\%$ when using different numbers of modules. Although partial FedDynAT and FedRBN can largely reduce the amount of computation, when training a large model that exceeds the memory limit, they need to repeatedly fetch and load small parts of the entire model from the cloud or the external storage during each forward and backward propagation. Since fetching and loading model parameters are usually much slower than forward and backward propagation, partial FedDynAT and FedRBN are far less efficient than they appear to be.

\subsection{Performance of FADE}\label{Subsec:performance}

\begin{figure*}[t]
\centering
\includegraphics[width=1.0\linewidth]{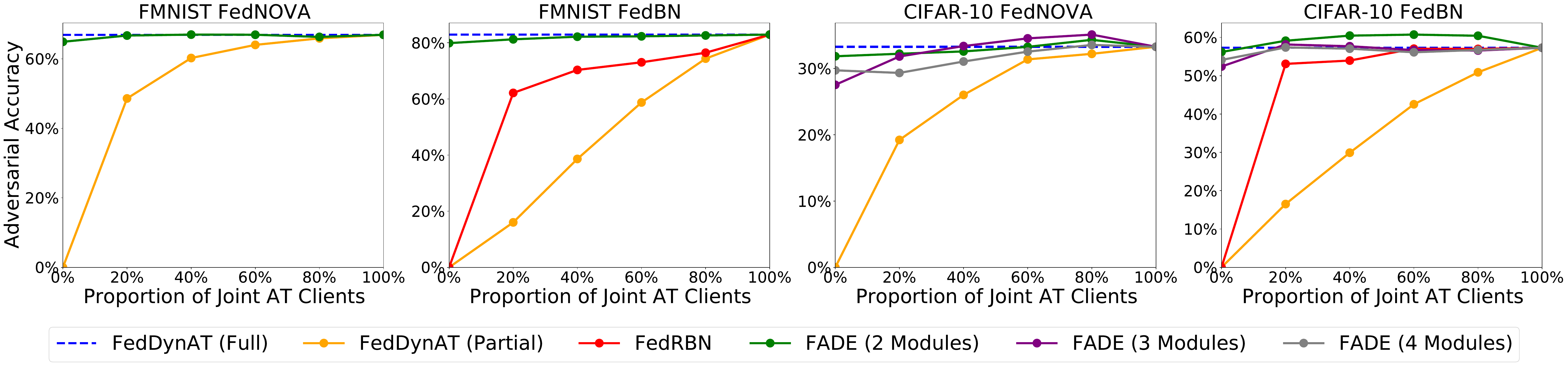}
\vspace{-2em}
\caption{Adversarial accuracy when training with different proportions of resource-sufficient clients who perform joint AT.}
\label{fig:frac}
\end{figure*}

We first compare FADE with other baselines in a setting where fixed $20\%$ clients can afford AT on the entire model, while the other $80\%$ clients can only afford ST on the entire model or AT on a single small module. \cref{table:fade_fmnist_res} and \cref{table:fade_cifar_res} show the natural and adversarial accuracy of different training schemes on FMNIST and CIFAR-10 respectively. For FADE, we mix clients using different numbers of modules with different ratios in each scheme. For example, ``FADE (2:2:3:3)'' means that the clients with 1 module, 2 modules, 3 modules and 4 modules are mixed in a ratio of 2:2:3:3. 

While neither partial FedDynAT nor FedRBN can maintain robustness under this resource constraint, the results show that FADE can attain almost the same or even higher accuracy and robustness compared to full FedDynAT (the constraint-free case). Additionally, the consistency in the performance of FADE with different mixes of clients shows the high compatibility of our flexible FDL framework.

We also conducted experiments with different proportions of resource-sufficient clients, as shown in \cref{fig:frac}. In each setting, the resource-sufficient clients who perform joint AT are mixed with resource-constrained clients who perform joint ST or adopt FADE with a multi-module model partition (e.g., a 2-module partition in ``FADE (2 Modules)''). 

We can see that even in the worst case that none of the clients have enough resources to complete AT on the entire model, FADE can achieve robustness comparable to full FedDynAT. And with only $40\%$ resource-sufficient clients, FADE can attain almost the same robustness as full FedDynAT in most experiments, while the other baselines still have significant robustness gaps from full FedDynAT under this setting.

It is noteworthy that the results of FADE with $0\%$ resource-sufficient clients in \cref{fig:frac} can be viewed as a special baseline: naively deploying DGL in FL without differentiated model partitions. We can see that the performance of this baseline is always worse than FADE with differentiated model partitions when the proportion of resource-sufficient clients is larger than $0\%$.

\subsection{The Influence of Auxiliary Weight Decay}

\begin{figure*}[t]
\centering
\includegraphics[width=1.0\linewidth]{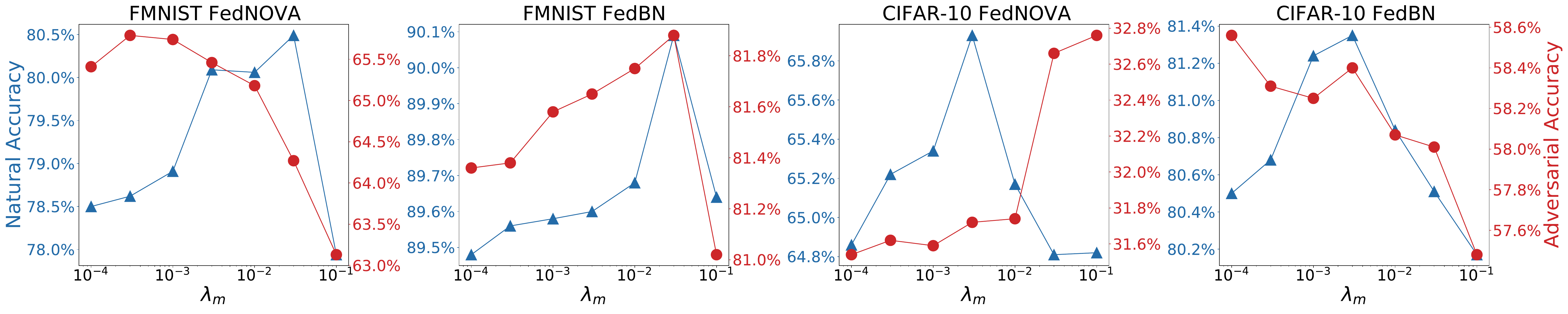}
\vspace{-2em}
\caption{Natural accuracy (blue lines with triangle markers) and adversarial accuracy (red lines with circle markers) of FADE (2:8) with different auxiliary weight decay hyperparameter $\lambda_m$.}
\label{fig:ablation}
\vspace{-0.5em}
\end{figure*}
As we discussed in \cref{Subsec:wd}, the hyperparameter $\lambda_m$ for weight decay on the auxiliary model acts as an important role that balances natural accuracy and adversarial robustness. To show the influence of this hyperparameter, we conduct experiments on ``FADE (2:8)'' (or ``FADE (2:8:0:0)'') with different $\lambda_m$ between $0.0001$ and $0.1$, and we plot the natural and adversarial accuracy in \cref{fig:ablation}.

We can observe that in all our settings, the natural accuracy increases first as we increase $\lambda_m$, and then goes down quickly. The growing part can be explained by our theory in \cref{Subsec:wd} that the large auxiliary weight decay can alleviate the objective inconsistency and improve the performance. However, when we adopt an excessively large weight decay, the weight decay will drive the model away from optimum and lead to a performance drop, which is also commonly observed in the joint training process.

For the adversarial accuracy, the effects of $\lambda_m$ become more complicated, since larger $\lambda_m$ can decrease both $g_m$ and $\mu_m$, which affect the robustness in opposite ways. An increasing adversarial accuracy suggests that the effect of $g_m$ dominates, while a decreasing one suggests that the effect of $\mu_m$ dominates. Similarly to the natural accuracy, we observe that the adversarial accuracy grows first and then goes down in most cases, which implies a stronger effect of $g_m$ when $\lambda_m$ is small. And considering the increasing natural accuracy, adopting a moderately large $\lambda_m$ usually attains a better overall performance on clean and adversarial examples.
\section{Related Works}
\textbf{Federated Learning.} Client-wise heterogeneity is one of the challenges that hinder the practice of Federated Learning (FL). Many studies have tried to overcome the statistical heterogeneity in data~\cite{karimireddy2019scaffold,liang2019variance,wang2020federated,tang2022fedcor} and the systematic heterogeneity in hardware~\cite{li2018federated,wang2020tackling,li2021hermes}. Beyond the heterogeneity, FL is also vulnerable to several kinds of attack, such as model poisoning attacks~\cite{bhagoji2019analyzing,sun2021fl} and adversarial examples~\cite{zizzo2020fat,shah2021adversarial}. In this paper, we mainly focus on the adversarial examples and deal with the challenge in federated adversarial training under client-wise heterogeneity~\cite{hong2021federated}.

\textbf{Adversarial Training.} AT is well known for its high demand for computing resources, including computing power and memory capacity~\cite{wong2020fast,liu2020adversarial}. Several fast AT algorithms have been proposed to reduce the computation in AT~\cite{shafahi2019adversarial,zhang2019you}, such as replacing PGD with FGSM~\cite{andriushchenko2020understanding,wong2020fast} or using other regularization methods for robustness~\cite{moosavi2019robustness,qin2019adversarial}. As a complement to these fast AT algorithms, FADE is proposed to reduce the memory requirement in AT, and we leave the combination of FADE and fast AT algorithms to future works.

\textbf{Decoupled Greedy Learning.} As deeper and deeper neural networks are used for better performance, the low efficiency of end-to-end (joint) training is exposed because it hinders the model parallelization and requires large memory for model parameters and intermediate results~\cite{hettinger2017forward,belilovsky2020decoupled}. As an alternative, Decoupled Greedy Learning (DGL) is proposed, which decouples the whole neural network into several modules and trains them separately without gradient dependency~\cite{marquez2018deep,belilovsky2019greedy,belilovsky2020decoupled,wang2021revisiting}. In contrast to allowing only one unified model partition in DGL, FADE fits better in FL with differentiated model partitions for heterogeneous clients. Additionally, FADE complements DGL with provable adversarial robustness and objective inconsistency alleviation, which allows FADE to maintain high adversarial and natural accuracy.
\section{Conclusions}
In this paper, we propose Federated Adversarial Decoupled Learning (FADE), a novel framework that enables federated adversarial training on heterogeneous resource-constrained edge devices. We theoretically analyze the convergence, robustness and objective inconsistency of FADE. Our experimental results reveal that FADE can significantly reduce both memory and computational requirements on small edge devices while maintaining almost the same accuracy and robustness as joint federated adversarial training.



\bibliography{ref}
\bibliographystyle{icml2023}

\newpage
\appendix
\onecolumn
\appendix

\allowdisplaybreaks[3]
\section{Convergence Analysis of Federated Decoupled Learning}\label{Apx:convergence}
\subsection{Preliminary}
In this section, we analyze the convergence property of Federated Decoupled Learning (FDL). Since FDL partitions the entire model with layers as the smallest unit, we only need to prove the convergence of each layer. We use $\boldsymbol{\omega}_n$ to denote all the parameters in layer $n$, and $m_k(n)$ to denote the module that contains layer $n$ on client $k$. We define the parameters other than $\boldsymbol{\omega}_n$ in module $m_k(n)$ as $\boldsymbol{\Omega}_{n,k}$. For notation simplicity, we also define $\vz_{n-1,k}=\vz_{m_k(n)-1}$, and the locally supervised loss of layer $n$ on client $k$ as:
\begin{align}
    l_{n,k}^{(t,j)}(\vz_{n-1}^{(t)},y;\boldsymbol{\omega}_n^{(t,j)})=l_{m_k(n)}(\vz_{m_k(n)-1}^{(t)},y;\boldsymbol{\omega}_n^{(t,j)},\boldsymbol{\Omega}_{n,k}^{(t,j)}),
\end{align}
where $l_{n,k}^{(t,j)}$ changes every iteration because of the update of $\boldsymbol{\Omega}_{n,k}^{(t,j)}$. For simplicity, from now on we abridge $(\vz,y)$ as $\vz$. We let $\vz_{n-1,k}^{(t,j)}$ follow the distribution with probability density $p_{n-1,k}^{(t)}(\rvz)$ at communication round $t$, and we define its converged density as $p_{n-1,k}^{*}(\rvz)$ with converged previous layers~\cite{belilovsky2020decoupled}. For some $\boldsymbol{\Omega}_{n,k}^*$, we define
\begin{align}
    &L_{n,k}^{(t)}(\boldsymbol{\omega}_n^{(t)})=\mathbb{E}_{\vz_{n-1,k}^{(t,j)}\sim p_{n-1,k}^{(t)}} \left[\frac{1}{\tau}\sum_{j=0}^{\tau-1}l_{n,k}^{(t,j)}(\vz_{n-1,k}^{(t,j)};\boldsymbol{\omega}_n^{(t)})\right];\label{eqlocal}\\
    &L_n^{(t)}(\boldsymbol{\omega}_n^{(t)})=\sum_{k=1}^N q_kL_{n,k}^{(t)}(\boldsymbol{\omega}_n^{(t)});\\
    &L_{n,k}(\boldsymbol{\omega}_n^{(t)})=\mathbb{E}_{\vz_{n-1,k}\sim p_{n-1,k}^{*}} \left[l_{n,k}^{*}(\vz_{n-1,k};\boldsymbol{\omega}_n^{(t)})\right]\nonumber\\
    &=\mathbb{E}_{\vz_{m_k(n)-1,k}\sim p_{m_k(n)-1,k}^{*}} \left[l_{m_k(n)}(\vz_{m_k(n)-1,k};\boldsymbol{\omega}_n^{(t)},\boldsymbol{\Omega}_{n,k}^{*})\right];\\
    &L_{n}(\boldsymbol{\omega}_n^{(t)})=\sum_{k=1}^N q_kL_{n,k}(\boldsymbol{\omega}_n^{(t)});
\end{align}
Following \citet{belilovsky2020decoupled}, we use the distance between the current density and the converged density below for our analysis:
\begin{align}
    \rho_{n}^{(t)} \triangleq \sum_{k=1}^N q_k\int\left|p_{n-1,k}^{(t)}(\rvz)-p_{n-1,k}^{*}(\rvz)\right| d\rvz,
\end{align}
And we also define the following gap between $l_{n,k}^{(t)}$ and $l_{n,k}^*$:
\begin{align}
    \xi_{n}^{(t)}\triangleq\sum_{k=1}^N\sum_{j=0}^{\tau-1}\frac{q_k}{\tau}\left\|\mathbb{E}_{\vz_{n-1,k}\sim p_{n-1}^*}\left[\nabla  l_{n,k}^{(t,j)}(\vz_{n-1,k};\omega_n^{(t)})-\nabla l_{n,k}^*(\vz_{n-1,k};\omega_n^{(t)})\right]\right\|^2.
\end{align}

We will discuss the convergence of $L_n(\boldsymbol{\omega}_n)$ for each layer $n$. Without specifying, all the gradients ($\nabla L$ or $\nabla l$) in the following analysis are with respect to $\boldsymbol{\omega}_n$. Following \citet{belilovsky2020decoupled} and \citet{wang2020tackling}, we make the common assumptions below. 

\begin{assumption}[$\mathcal{L}$-smoothness~\cite{belilovsky2020decoupled,wang2020tackling}] \label{ass1}
$L_n$ is differentiable with respect to $\boldsymbol{\omega}_n^{(t)}$ and its gradient is $\mathcal{L}_n$-Lipschitz for all $t$. Similarly, $l_{n,k}^{(t,j)}$ is differentiable with respect to $\boldsymbol{\omega}_{n,k}$ and its gradient is $\widetilde{\mathcal{L}}_{n}$-Lipschitz for all $t,j$ and $k$.
\end{assumption}

\begin{assumption}[Robbins-Monro conditions~\cite{belilovsky2020decoupled}] \label{ass2}
The learning rates satisfy $\sum_{t=0}^\infty \eta_{t}=\infty$ yet $\sum_{t=0}^\infty \eta_{t}^{2}<\infty$.
\end{assumption}

\begin{assumption}[Finite variance~\cite{belilovsky2020decoupled}] \label{ass3}
There exists some positive constant $G$ such that $\forall t,j$ and $\forall k$, $\mathbb{E}_{\vz_{n-1,k}^{(t,j)}\sim p_{n-1,k}^{(t,j)}}\left[\left\|\nabla l_{n,k}^{(t,j)}(\vz_{n-1,k}^{(t,j)};\boldsymbol{\omega}_n)\right\|^{2}\right] \leq G$ and  $\mathbb{E}_{\vz_{n-1,k}\sim p_{n-1}^{*}}\left[\left\|\nabla l_{n,k}^*(\vz_{n-1,k};\boldsymbol{\omega}_n)\right\|^{2}\right] \leq G$ at any $\boldsymbol{\omega}_n$.
\end{assumption}


\begin{assumption}[Convergence of the previous modules and $\boldsymbol{\Omega}_{n,k}$~\cite{belilovsky2020decoupled}] \label{ass5}
We assume that $\sum_{t=0}^\infty \rho_{n}^{(t)}<\infty$ and $\sum_{t=0}^\infty {\xi_{n}^{(t)}}<\infty$.
\end{assumption}

\subsection{Proof of \cref{theorem:convergence}}
With all above assumptions, we get the following theorem that guarantees the convergence of Federated Decoupled Learning. 

\begin{theorem}
Under \cref{ass1} - \cref{ass5}, given a client sampling method that satisfies $\Pr(k\in \sS_n^{(t)})=p$ for any client $k$, Federated Decoupled Learning converges as follows: 
\begin{align}
&\inf_{t \leq T} \mathbb{E}\left[\left\|\nabla L_{n}\left(\boldsymbol{\omega}_n^{(t)}\right)\right\|^{2}\right] \nonumber\\
\leq& \mathcal{O}\left(\frac{1}{\sum_{t=0}^{T} \eta_{t}}\right)+\mathcal{O}\left(\frac{\sum_{t=0}^{T} {\rho_n^{(t)}} \eta_{t}}{\sum_{t=0}^{T} \eta_{t}}\right)+\mathcal{O}\left(\frac{\sum_{t=0}^{T} {\xi_n^{(t)}} \eta_{t}}{\sum_{t=0}^{T} \eta_{t}}\right)+\mathcal{O}\left(\frac{\sum_{t=0}^{T} \eta_{t}^2}{\sum_{t=0}^{T} \eta_{t}}\right).
\end{align}

\begin{proof}
We consider the following SGD scheme with learning rate $\left\{\eta_{t}\right\}_{t}$:
\begin{align}
\boldsymbol{\omega}_n^{(t+1)}=\boldsymbol{\omega}_n^{(t)}-\eta_t\tau\frac{\sum_{k\in\sS_n^{(t)}}q_k\boldsymbol{h}_{n,k}^{(t)}}{p},\label{eq:proxy_update}
\end{align}
where $\sS_{n}^{(t)}=\{k\in\sS^{(t)}:n \in m_k^t\}$ which is defined in \cref{eq:aggre}. And ${\boldsymbol{h}_{n,k}^{(t)}}$ is defined as 
\begin{align*}
    \boldsymbol{h}_{n,k}^{(t)}= \frac{1}{\tau}\sum_{j=0}^{\tau-1} \nabla l_{n,k}^{(t,j)}(\vz_{n-1,k}^{(t,j)};\boldsymbol{\omega}_{n,k}^{(t,j)}).
\end{align*}
According to the Lipschitz-smooth assumption for the global objective function $L_{n}$, it follows that 
\begin{align}
& \mathbb{E}\left[L_{n}\left(\boldsymbol{\omega}_n^{(t+1)}\right)\right]-L_{n}\left(\boldsymbol{\omega}_n^{(t)}\right) \nonumber\\
\leq &- \eta_t\tau \underbrace{\mathbb{E}\left[\left\langle\nabla L_{n}\left(\boldsymbol{\omega}_n^{(t)}\right), \frac{\sum_{k\in\sS_n^{(t)}}q_k\boldsymbol{h}_{n,k}^{(t)}}{p}\right\rangle\right]}_{T_{1}}+\frac{\eta_t^{2}\tau^2 \mathcal{L}_n}{2p^2} \underbrace{\mathbb{E}\left[\left\|\sum_{k\in\sS_n^{(t)}}q_k\boldsymbol{h}_{n,k}^{(t)}\right\|^{2}\right]}_{T_{2}},
\label{proof:T1T2}
\end{align}
where expectation is taken over the minibatch as well as $\sS_n^{(t)}$.

Similar to the proof in \cite{wang2020tackling}, to bound the $T_1$ in Inequality~(\ref{proof:T1T2}), we should notice that
\begin{align}
T_{1} = &\mathbb{E}\left[\left\langle\nabla L_{n}\left(\boldsymbol{\omega}_n^{(t)}\right), \frac{\sum_{k\in\sS_n^{(t)}}q_k\boldsymbol{h}_{n,k}^{(t)}}{p}\right\rangle\right]\nonumber \\
= & \left\langle\nabla L_{n}\left(\boldsymbol{\omega}_n^{(t)}\right), \frac{\mathbb E\left[\sum_{k\in\sS_n^{(t)}}q_k\boldsymbol{h}_{n,k}^{(t)}\right]}{p}\right\rangle\nonumber \\
= & \left\langle\nabla L_{n}\left(\boldsymbol{\omega}_n^{(t)}\right), \frac{\sum_{k=1}^N\mathbb E\left[q_k\boldsymbol{h}_{n,k}^{(t)}1_{[k\in \sS_n^{(t)}]}\right]}{p}\right\rangle\nonumber \\
= &\left\langle\nabla L_{n}\left(\boldsymbol{\omega}_n^{(t)}\right), {\sum_{k=1}^{N} q_{k}\mathbb{E}\boldsymbol{h}_{n,k}^{(t)}}\right\rangle \nonumber\\
=&\frac{1}{2}\left\|\nabla L_{n}\left(\boldsymbol{\omega}_n^{(t)}\right)\right\|^{2}+\frac{1}{2} \left\|{\sum_{k=1}^{N} q_{k}\mathbb{E}\boldsymbol{h}_{n,k}^{(t)}}\right\|^2-\frac{1}{2} \left\|\nabla L_{n}\left(\boldsymbol{\omega}_n^{(t)}\right)-{\sum_{k=1}^{N} q_{k}\mathbb{E}\boldsymbol{h}_{n,k}^{(t)}}\right\|^{2} \label{firseq}\\
\geq &\frac{1}{2}\left\|\nabla L_{n}\left(\boldsymbol{\omega}_n^{(t)}\right)\right\|^{2} -\frac{1}{2} \left\|\nabla L_{n}\left(\boldsymbol{\omega}_n^{(t)}\right)-{\sum_{k=1}^{N} q_{k}\mathbb{E}\boldsymbol{h}_{n,k}^{(t)}}\right\|^{2}\nonumber \\
\geq &\frac{1}{2}\left\|\nabla L_{n}\left(\boldsymbol{\omega}_n^{(t)}\right)\right\|^{2} - \left\|\nabla L_n^{(t)}\left(\boldsymbol{\omega}_n^{(t)}\right)-{\sum_{k=1}^{N} q_{k}\mathbb{E}\boldsymbol{h}_{n,k}^{(t)}}\right\|^{2}
-\left\|\nabla L_{n}\left(\boldsymbol{\omega}_n^{(t)}\right)-\nabla L_n^{(t)}(\boldsymbol{\omega}_n^{(t)})\right\|^{2} \label{secondeq} \\
\geq &\frac{1}{2}\left\|\nabla L_{n}\left(\boldsymbol{\omega}_n^{(t)}\right)\right\|^{2}- {\sum_{k=1}^{N} q_{k}\left\|\mathbb{E}\boldsymbol{h}_{n,k}^{(t)}-\nabla L_{n,k}^{(t)}(\boldsymbol{\omega}_n^{(t)})\right\|^{2}}-\left\|\nabla L_{n}\left(\boldsymbol{\omega}_n^{(t)}\right)-\nabla L_n^{(t)}(\boldsymbol{\omega}_n^{(t)})\right\|^{2}. \label{thirdeq}
\end{align}
\cref{firseq} uses the fact: $2\langle a, b\rangle=\|a\|^{2}+\|b\|^{2}-\|a-b\|^{2}$, and Inequality~(\ref{secondeq}) uses the fact: $\|a+b\|^2\leq 2\|a\|^2+2\|b\|^2$. Inequality~(\ref{thirdeq}) uses $L_n^{(t)}={\sum_{k=1}^N q_kL_{n,k}^{(t)} }$ and Jenson's inequality $\left\|\sum_{i=k}^{N} q_{k} \va_{i}\right\|^{2} \leq \sum_{k=1}^{N} q_{k}\left\|\va_{i}\right\|^{2}$.

For the second term in $T_1$, we have 
\begin{align*}
\left\|\mathbb{E}\boldsymbol{h}_{n,k}^{(t)}-\nabla L_{n,k}^{(t)}(\boldsymbol{\omega}_n^{(t)})\right\|^2&=\left\|\frac{1}{\tau}\sum_{j=0}^{\tau-1}\mathbb E\left[\nabla l_{n,k}^{(t,j)}(\vz_{n-1,k}^{(t,j)};\boldsymbol{\omega}_{n,k}^{(t,j)})-\nabla l_{n,k}^{(t,j)}(\vz_{n-1,k}^{(t,j)};\boldsymbol{\omega}_n^{(t,0)})\right]\right\|^2\\
&\le\frac{1}{\tau}\sum_{j=0}^{\tau-1}\mathbb E\left\|\nabla l_{n,k}^{(t,j)}(\vz_{n-1,k}^{(t,j)};\boldsymbol{\omega}_{n,k}^{(t,j)})-\nabla l_{n,k}^{(t,j)}(\vz_{n-1,k}^{(t,j)};\boldsymbol{\omega}_n^{(t,0)})\right\|^2\\
&\le \frac{\mathcal{\widetilde L}_n^2}{\tau}\sum_{j=0}^{\tau-1}\mathbb E\left\|\boldsymbol{\omega}_{n,k}^{(t,j)}-\boldsymbol{\omega}_{n,k}^{(t,0)}\right\|^2.
\end{align*}

And we know that
\begin{align*}
    \mathbb E\left\|\boldsymbol{\omega}_{n,k}^{(t,j)}-\boldsymbol{\omega}_{n,k}^{(t,0)}\right\|^2 &= \eta_t^2\mathbb E\left\|\sum_{s=0}^{j-1}\nabla l_{n,k}^{(t,s)}\right\|^2\\
    &\le \eta_t^2j\sum_{s=0}^{j-1}\mathbb E\left\|\nabla l_{n,k}^{(t,s)}\right\|^2\\
    &\le \eta_t^2 j^2G.
\end{align*}

Thus we get
\begin{align}
    \left\|\mathbb{E}\boldsymbol{h}_{n,k}^{(t)}-\nabla L_{n,k}^{(t)}(\boldsymbol{\omega}_n^{(t)})\right\|^2&\le \frac{\mathcal{\widetilde L}_n^2}{\tau}\sum_{j=0}^{\tau-1}\mathbb E\left\|\boldsymbol{\omega}_{n,k}^{(t,j)}-\boldsymbol{\omega}_{n,k}^{(t,0)}\right\|^2\nonumber\\
    &\le \frac{\mathcal{\widetilde L}_n^2}{\tau}\sum_{j=0}^{\tau-1}\eta_t^2 j^2G\nonumber\\
    &\le \frac{\tau^2}{2}\mathcal{\widetilde L}_n^2\eta_t^2G.\label{eq:T1_second}
\end{align}

For the third term in $T_1$, based on the proof of Lemma 3.2 in \citet{belilovsky2020decoupled}, we have 
\begin{align}
&\left\|\nabla L_{n}\left(\boldsymbol{\omega}_n^{(t)}\right)-\nabla L_n^{(t)}\left(\boldsymbol{\omega}_n^{(t)}\right)\right\|^{2} \nonumber\\
=& \left\|\sum_k q_k\mathbb{E}_{\vz\sim p_{n-1,k}^{*}} \left[\nabla l_{n,k}^*(\vz;\boldsymbol{\omega}_n^{(t)})\right]-\sum_k q_k\mathbb{E}_{\vz^{(t,j)}\sim p_{n-1,k}^{(t)}} \left[\frac{1}{\tau}\sum_{j}\nabla l_{n,k}^{(t,j)}(\vz^{(t,j)};\boldsymbol{\omega}_n^{(t)})\right]\right\|^2\nonumber\\
=&\left\|\frac{1}{\tau}\sum_k q_k\sum_j\left(\mathbb{E}_{\vz\sim p_{n-1,k}^{*}} \left[\nabla l_{n,k}^*(\vz;\boldsymbol{\omega}_n^{(t)})\right]-\mathbb{E}_{\vz\sim p_{n-1,k}^{(t)}} \left[\nabla l_{n,k}^{(t,j)}(\vz;\boldsymbol{\omega}_n^{(t)})\right]\right)\right\|^2\nonumber\\
\le&\frac{1}{\tau}\sum_k q_k\sum_j\left\|\mathbb{E}_{\vz\sim p_{n-1,k}^{*}} \left[\nabla l_{n,k}^*(\vz;\boldsymbol{\omega}_n^{(t)})\right]-\mathbb{E}_{\vz\sim p_{n-1,k}^{(t)}} \left[\nabla l_{n,k}^{(t,j)}(\vz;\boldsymbol{\omega}_n^{(t)})\right]\right\|^2\nonumber\\
\le&\frac{2}{\tau}\sum_k q_k\sum_j\left\|\int \nabla l_{n,k}^{(t,j)}(\vz;\boldsymbol{\omega}_n^{(t)})p_{n-1,k}^{(t)}(\rvz)d\rvz -\int \nabla l_{n,k}^{(t,j)}(\vz;\boldsymbol{\omega}_n^{(t)})p_{n-1,k}^{*}(\rvz)d\rvz \right\|^2 \nonumber\\
&+\frac{2}{\tau}\sum_k q_k\sum_j\left\| \mathbb{E}_{\vz\sim p_{n-1,k}^{*}}\left[\nabla l_{n,k}^{(t,j)}(\vz;\boldsymbol{\omega}_n^{(t)})-\nabla l_{n,k}^{*}(\vz;\boldsymbol{\omega}_n^{(t)})\right]\ \right\|^2\nonumber\\
\le&  \frac{2}{\tau}\sum_k q_k\sum_j\left(\int \left\|\nabla l_{n,k}^{(t,j)}(\rvz;\boldsymbol{\omega}_n^{(t)})\right\|\sqrt{|p_{n-1,k}^{(t)}(\rvz)-p_{n-1,k}^{*}(\rvz)| |p_{n-1,k}^{(t)}(\rvz)-p_{n-1,k}^{*}(\rvz) |} d\rvz\right)^2\nonumber\\&+2\xi_n^{(t)}\nonumber\\
\le&\frac{2}{\tau}\sum_k q_k\sum_j\int \left\|\nabla l_{n,k}^{(t,j)}(\rvz;\boldsymbol{\omega}_n^{(t)})\right\|^2{ |p_{n-1,k}^{(t)}(\rvz)-p_{n-1}^{*}(\rvz) | } d\rvz\int|p_{n-1,k}^{(t)}(\rvz)-p_{n-1,k}^{*}(\rvz)|d\rvz\nonumber\\&+2\xi_n^{(t)}\nonumber\\
\le &\frac{2}{\tau} \sum_kq_k\int|p_{n-1,k}^{(t)}(\rvz)-p_{n-1,k}^{*}(\rvz)|d\rvz\sum_j \int \left\|\nabla l_{n,k}^{(t,j)}(\rvz;\boldsymbol{\omega}_n^{(t)})\right\|^2{ \left(p_{n-1,k}^{(t)}(\rvz)+p_{n-1,k}^{*}(\rvz) \right) } d\rvz\nonumber\\&+2\xi_n^{(t)}\nonumber\\
\le &4G\rho_n^{(t)}+2\xi_n^{(t)}.\label{eq:T1_third}
\end{align}

Plugging \cref{eq:T1_second} and \cref{eq:T1_third} back to \cref{thirdeq}, we have 
\begin{align}
T_{1} & \geq \frac{1}{2}\left\|\nabla L_{n}\left(\boldsymbol{\omega}_n^{(t)}\right)\right\|^{2}- \frac{\tau^2}{2}\mathcal{\widetilde L}_n^2\eta_t^2G-4G\rho_n^{(t)}-2\xi_n^{(t)}.\label{eq:T1_bound}
\end{align}

Now we turn to bound $T_2$. With the fact that $\sum_{k\in \sS_n^{(t)}}q_k\le 1$, we have:
\begin{align} 
T_{2} &=\mathbb{E}\left[\left(\sum_{k\in \sS_n^{(t)}}q_k\right)^2\left\| \frac{\sum_{k\in\sS_n^{(t)}}q_k\boldsymbol{h}_{n,k}^{(t)}}{\sum_{k\in \sS_n^{(t)}}q_k}\right\|^{2}\right]\nonumber\\
&\le \mathbb{E}\left[\sum_{k\in\sS_n^{(t)}}q_k\left\|\boldsymbol{h}_{n,k}^{(t)}\right\|^{2}\right]\nonumber\\
&\le \mathbb{E}\left[ \frac{\sum_{k\in\sS_n^{(t)}}q_k\sum_{j=0}^{\tau-1}\left\|\nabla l_{n,k}^{(t,j)}(\vz_{n-1,k}^{(t,j)};\boldsymbol{\omega}_{n,k}^{(t,j)})\right\|^{2}}{\tau}\right]\nonumber\\
&= \frac{\sum_{k=1}^Nq_k\sum_{j=0}^{\tau-1}\mathbb E\left\|\nabla l_{n,k}^{(t,j)}(\vz_{n-1,k}^{(t,j)};\boldsymbol{\omega}_{n,k}^{(t,j)})\right\|^2\mathbb E_{\sS_n^{(t)}}\left[1_{k\in\sS_n^{(t)}}\right]}{\tau}\nonumber\\
&\le pG.\label{eq:T2_bound}
\end{align}

Instituting $T_1$ and $T_2$ in \cref{proof:T1T2} with \cref{eq:T1_bound} and \cref{eq:T2_bound} respectively, we have
\begin{align}
    &\mathbb{E}\left[L_{n}\left(\boldsymbol{\omega}_n^{(t+1)}\right)\right]-L_{n}\left(\boldsymbol{\omega}_n^{(t)}\right)\nonumber\\
    \le &-\eta_t\tau\left[\frac{1}{2}\left\|\nabla L_{n}\left(\boldsymbol{\omega}_n^{(t)}\right)\right\|^{2}- \frac{\tau^2}{2}\mathcal{\widetilde L}_n^2\eta_t^2G-4G\rho_n^{(t)}-2\xi_n^{(t)}\right]+\frac{\eta_t^{2}\tau^2 \mathcal{L}_nG}{2p}.\label{eq:raw_diff}
\end{align}

Assuming that $\eta_t\le1$ and $L_n>0$, rearranging \cref{eq:raw_diff}, taking the expectation and averaging across all rounds, one can obtain
\begin{align*}
    &\frac{1}{2}\sum_{t=0}^T \eta_t \mathbb{E}\left[\left\|\nabla L_{n}\left(\boldsymbol{\omega}_n^{(t)}\right)\right\|^{2}\right]\\ 
    \leq& \frac{1}{\tau}\left(L_{n}\left(\boldsymbol{\omega}_n^{(0)}\right)-\mathbb E\left[L_{n}\left(\boldsymbol{\omega}_n^{(T+1)}\right)\right]\right)+A\sum_{t=0}^T\eta_t^2+B\sum_{t=0}^T\eta_t\rho_n^{(t)}+C\sum_{t=0}^T\eta_t\xi_n^{(t)}\\
    \leq& \frac{1}{\tau}L_{n}\left(\boldsymbol{\omega}_n^{(0)}\right)+A\sum_{t=0}^T\eta_t^2+B\sum_{t=0}^T\eta_t\rho_n^{(t)}+C\sum_{t=0}^T\eta_t\xi_n^{(t)},
\end{align*}
where $A,B$ and $C$ are some positive constants. Now we get our final result:
\begin{align*}
    &\inf _{t \leq T} \mathbb{E}\left[\left\|\nabla L_{n}\left(\boldsymbol{\omega}_n^{(t)}\right)\right\|^{2}\right]\leq \frac{1}{\sum_{t=0}^T \eta_t}\sum_{t=0}^T \eta_t \mathbb{E}\left[\left\|\nabla L_{n}\left(\boldsymbol{\omega}_n^{(t)}\right)\right\|^{2}\right]\\
    \leq& \mathcal{O}\left(\frac{1}{\sum_{t=0}^{T} \eta_{t}}\right)+
\mathcal{O}\left(\frac{\sum_{t=0}^{T} {\rho_n^{(t)}} \eta_{t}}{\sum_{t=0}^{T} \eta_{t}}\right)+\mathcal{O}\left(\frac{\sum_{t=0}^{T} {\xi_n^{(t)}} \eta_{t}}{\sum_{t=0}^{T} \eta_{t}}\right)+\mathcal{O}\left(\frac{\sum_{t=0}^{T} \eta_{t}^2}{\sum_{t=0}^{T} \eta_{t}}\right).
\end{align*}

It is simple to verify that $\frac{1}{\sum_{t=0}^{T} \eta_{t}} \to 0$ and $\frac{\sum_{t=0}^{T} \eta_{t}^2}{\sum_{t=0}^{T} \eta_{t}} \to 0$ if $T\to \infty$. As for $\frac{\sum_{t=0}^{T} {\rho_n^{(t)}} \eta_{t}}{\sum_{t=0}^{T} \eta_{t}}$, according to the Cauchy-Schwartz inequality, we have 
\begin{align*}
    {\sum_{t=0}^{T} {\rho_n^{(t)}} \eta_{t}}&=\sum_{t=0}^{T} \sqrt{\rho_n^{(t)}} \left(\sqrt{\rho_n^{(t)}} \eta_{t}\right)\\
    &\leq \sqrt{\left(\sum_{t=0}^{T} {\rho_n^{(t)}} \right)\left(\sum_{t=0}^{T} {\rho_n^{(t)}}\eta_t^2\right)}\\
    & \leq \sqrt{\left(\sum_{t=0}^{T} {\rho_n^{(t)}} \right)\left(\sum_{t=0}^{T} {\rho_n^{(t)}}\right)\left(\sum_{t=0}^{T} \eta_t^2\right)} \\
    &< \infty.
\end{align*}
Hence, we also have $\frac{\sum_{t=0}^{T} {\rho_n^{(t)}} \eta_{t}}{\sum_{t=0}^{T} \eta_{t}}\to 0$ if $T\to \infty$. Similarly, we get the same result for $\frac{\sum_{t=0}^{T} {\xi_n^{(t)}} \eta_{t}}{\sum_{t=0}^{T} \eta_{t}}$. In conclusion, we get the result in \cref{Subsec:fdl}:
\begin{align*}
\lim_{T\rightarrow \infty}\inf_{t \leq T} \mathbb{E}\left\|\nabla L_n\left(\boldsymbol{\omega}_n^{(t)}\right)\right\|^{2} = 0.
\end{align*}

We notice that there is a minor difference between the update rule in \cref{eq:aggre} and in \cref{eq:proxy_update}, where we can bridge the gap by setting the original learning rate in \cref{eq:aggre} as $\eta_t=\eta_t\sum_{i\in\sS_n^{(t)}}q_k/p$. 

\end{proof}

\end{theorem}

\section{Proofs of \cref{theorem:robustness}, \cref{theorem:inconsistency} and Additional Analysis}\label{Apx:proofs}
\subsection{Proof of \cref{theorem:robustness}}\label{pf:theorem:robustness}
\begin{theorem}
Assume that $\tilde l_m(\vz_m,y)$ is $\mu_m$-strongly convex in $\vz_m$ for each module $m$. We can guarantee that the entire model has a joint $(\epsilon_0,c_M)$-robustness in $l(\vx,y)$, if each module $m\le M$ has local $(\epsilon_{m-1},c_m)$-robustness in $l_m(\vz_{m-1},y)$, and
\begin{align}
    \quad\epsilon_{m}\ge \frac{g_m}{\mu_m}+\sqrt{\frac{2c_m}{\mu_m}+\frac{g_m^2}{\mu_m^2}},
\end{align}
where $g_m=\left\|\nabla_{\vz_m}\tilde l_m(\vz_m,y)\right\|$.
\begin{proof}
We only need to prove the joint robustness of the concatenation of module $m$ and $(m+1)$ given the local robustness of them separately, and then we can use deduction to get the joint robustness of the entire model given the local robustness of all modules.

For a module $m$ and any perturbation $\boldsymbol{\delta}_{m-1}\in\{\boldsymbol{\delta}_{m-1}:\Vert \boldsymbol{\delta}_{m-1}\Vert \le \epsilon_{m-1}\}$ at its input, let $\vr=f_m(\vz_{m-1}+\boldsymbol{\delta}_{m-1})-f_m(\vz_{m-1})$. Given $\mu_m$-strongly convexity and $(\epsilon_{m-1},c_m)$-robustness in $\tilde l_m(\vz_m,y)$, we have
\begin{align}
    &\nabla_{\vz_m}\tilde l_m(\vz_m,y)^T\vr+\frac{\mu_m}{2}\Vert r\Vert^2\le \tilde l_m(\vz_m+\vr,y)-\tilde l_m(\vz_m,y)\le c_m\\
    \Rightarrow& \frac{\mu_m}{2}\left[\left\Vert\vr+\frac{\nabla_{\vz_m}\tilde l_m(\vz_m,y)}{\mu_m}\right\Vert^2-\frac{\Vert \nabla_{\vz_m}\tilde l_m(\vz_m,y)\Vert^2}{\mu_m^2}\right]\le c_m\\
    \Rightarrow& \left\Vert\vr+\frac{\nabla_{\vz_m}\tilde l_m(\vz_m,y)}{\mu_m} \right\Vert\le \sqrt{\frac{2c_m}{\mu_m}+\frac{\Vert \nabla_{\vz_m}\tilde l_m(\vz_m,y)\Vert^2}{\mu_m^2}}\\
    \Rightarrow& \Vert \vr \Vert \le \frac{\Vert\nabla_{\vz_m}\tilde l_m(\vz_m,y)\Vert}{\mu_m}+\sqrt{\frac{2c_m}{\mu_m}+\frac{\Vert \nabla_{\vz_m}\tilde l_m(\vz_m,y)\Vert^2}{\mu_m^2}}=\frac{g_m}{\mu_m}+\sqrt{\frac{2c_m}{\mu_m}+\frac{g_m^2}{\mu_m^2}}.
\end{align}
And we know that
\begin{align}
    \epsilon_m\ge \frac{g_m}{\mu_m}+\sqrt{\frac{2c_m}{\mu_m}+\frac{g_m^2}{\mu_m^2}}\ge \Vert\vr\Vert,
\end{align}
which gives us
\begin{align}
    \forall \boldsymbol{\delta}_{m-1}\in\{\boldsymbol{\delta}_{m-1}:\Vert \boldsymbol{\delta}_{m-1}\Vert \le \epsilon_{m-1}\}, \Vert f_m(\vz_{m-1}+\boldsymbol{\delta}_{m-1})-f_m(\vz_{m-1})\Vert \le \epsilon_m.
\end{align}
With the $(\epsilon_m,c_{m+1})$-robustness of module $(m+1)$, we have the joint robustness of the concatenation of $m$ and $(m+1)$:
\begin{gather}
    \forall \boldsymbol{\delta}_{m-1}\in\{\boldsymbol{\delta}_{m-1}:\Vert \boldsymbol{\delta}_{m-1}\Vert \le \epsilon_{m-1}\},\nonumber\\
    l_{m+1}(f_m(\vz_{m-1}+\boldsymbol{\delta}_{m-1}),y)-l_{m+1}(f_m(\vz_{m-1}),y)\le c_{m+1}.
\end{gather}
\end{proof}
\end{theorem}

\subsection{Proof of \cref{theorem:inconsistency}}\label{pf:theorem:inconsistency}
\begin{lemma}
Assume that $\tilde l_m(\vz_m,y)$ and $\tilde l'_m(\vz_{m},y)$ are $\beta_m,\beta'_{m}$-smooth in $\vz_m$ for a module $m$. If there exist $\tilde c_m$, $\tilde c'_{m}$, and $r\ge \sqrt{2\frac{\tilde c_m+\tilde c'_{m}}{\beta_m+\beta'_{m}}}$, such that the auxiliary model has $(r,\tilde c_m)$-robustness in $\tilde l_m(\vz_m,y)$, and the backbone network has $(r,\tilde c'_m)$-robustness in $\tilde l'_m(\vz_m,y)$, then we have:
\begin{align}
    \Vert\nabla_{\vw_m} l-\nabla_{\vw_m} l_m\Vert \le \left\Vert \frac{\partial \vz_m}{\partial \vw_m}\right\Vert\sqrt{2(\tilde c_m+\tilde c'_{m})(\beta_m+\beta'_{m})}.
\end{align}
\begin{proof}
With the chain rule, we know that
\begin{align}
    \nabla_{\vw_m} l-\nabla_{\vw_m} l_m=\frac{\partial \vz_m}{\partial \vw_m}\frac{\partial (l-l_m)}{\partial \vz_m}=\frac{\partial \vz_m}{\partial \vw_m}\frac{\partial (\tilde l'_m-\tilde l_m)}{\partial \vz_m},
\end{align}
and thus
\begin{align}
    \Vert\nabla_{\vw_m} l-\nabla_{\vw_m} l_m\Vert \le \left\Vert \frac{\partial \vz_m}{\partial \vw_m}\right\Vert\left\Vert\frac{\partial (\tilde l'_m-\tilde l_m)}{\partial \vz_m}\right\Vert.
\end{align}
We now need to find the upper bound of the second factor. We define $h(\vz_m)=\tilde l'_m(\vz_m,y)-\tilde l_m(\vz_m,y)$, which is $(\beta_m+\beta_m')$-smooth in $\vz_m$. For any $\boldsymbol{\delta}_m,\Vert \boldsymbol{\delta}_m\Vert\le r$, with the $(r,\tilde c_m)$-robustness in $\tilde l_m(\vz_m,y)$ and $(r,\tilde c_m')$-robustness in $\tilde l'_m(\vz_m,y)$, we have
\begin{align}
    \vert h(\vz_m+\boldsymbol{\delta}_m)-h(\vz_m)\vert &\le \vert \tilde l_m(\vz_m+\boldsymbol{\delta}_m,y)-\tilde l_m(\vz_m,y)\vert+\vert \tilde l'_m(\vz_m+\boldsymbol{\delta}_m,y)-\tilde l'_m(\vz_m,y)\vert\nonumber\\
    &\le \tilde c_m+\tilde c_m'.
\end{align}
And with the $(\beta_m+\beta_m')$-smoothness, we know that
\begin{align}
    \left(\frac{\partial h(\vz_m)}{\partial \vz_m}\right)^T\boldsymbol{\delta}_m-\frac{\beta_m+\beta_m'}{2}\Vert \boldsymbol{\delta}_m\Vert^2\le h(\vz_m+\boldsymbol{\delta}_m)-h(\vz_m)\le \tilde c_m+\tilde c_m'.
\end{align}
The maximum of the LHS is achieved when $\boldsymbol{\delta}^*_m=\frac{1}{\beta_m+\beta_m'}\frac{\partial h(\vz_m)}{\partial \vz_m}$, and thus we get
\begin{align}
    &\frac{1}{2(\beta_m+\beta_m')}\left\Vert\frac{\partial h(\vz_m)}{\partial \vz_m}\right\Vert^2\le \tilde c_m+\tilde c_m'\\
    \Rightarrow& \left\Vert\frac{\partial h(\vz_m)}{\partial \vz_m}\right\Vert\le \sqrt{2(\tilde c_m+\tilde c_m')(\beta_m+\beta_m')}.
\end{align}
To check the achievability of this maximum, we have
\begin{align}
    \Vert\boldsymbol{\delta}^*_m\Vert=\frac{1}{\beta_m+\beta_m'}\left\Vert\frac{\partial h(\vz_m)}{\partial \vz_m}\right\Vert\le \sqrt{2\frac{\tilde c_m+\tilde c'_{m}}{\beta_m+\beta'_{m}}}\le r.
\end{align}
Thus, we get our final result
\begin{align}
    \Vert\nabla_{\vw_m} l-\nabla_{\vw_m} l_m\Vert \le \left\Vert \frac{\partial \vz_m}{\partial \vw_m}\right\Vert\sqrt{2(\tilde c_m+\tilde c'_{m})(\beta_m+\beta'_{m})}.
\end{align}
\end{proof}
\end{lemma}

\subsection{Case Study: Linear Auxiliary Output Model}\label{Apx:case}
For a linear auxiliary output model $\boldsymbol{\theta}_m=\{\mW_m,\vb_m\}$, the cross-entropy loss is given as
\begin{align}
    \tilde l(\vz_m,\vy)=\mathcal{L}(\sigma(\mW_m^T\vz_m+\vb_m),\vy),
\end{align}
where $\mathcal{L}(\vp,\vy)=-\sum_{i=1} y_i\log(p_i)$ and $\sigma(\vq)_i = \exp(q_i)/(\sum_j \exp(q_j))$ are cross-entropy loss and softmax function respectively. Let $\vp_m =\sigma(\mW_m^T\vz_m+\vb_m)$, we know that
\begin{align}
    \nabla_{\vz_m}\tilde l(\vz_m,\vy)=\mW_m(\vp_m-\vy),
\end{align}
and 
\begin{align}
    \mH_m=\nabla^2_{\vz_m}\tilde l(\vz_m,\vy)=\mW_m\mJ_m\mW_m^T,
\end{align}
where
\begin{align}
    \mJ_m=\text{diag}(\vp_m)-\vp_m\vp_m^T
\end{align}
is the Jacobian of the softmax function. We have the following properties related to the robustness and objective consistency in \cref{theorem:robustness} and \cref{theorem:inconsistency}:

1. (First Order Property) Smaller $\Vert \mW_m\Vert$ leads to smaller $g_m$ and $\tilde c_m$.
\begin{align}
    g_m&=\Vert \nabla_{\vz_m}\tilde l_m(\vz_m,y)\Vert=\Vert \mW_m(\vp_m-\vy)\Vert \le \sqrt{2}\Vert \mW_m\Vert,\\
    \tilde c_m&=\max_{\Vert\boldsymbol{\delta}_m\Vert\le r}\vert \tilde l_m(\vz_m+\boldsymbol{\delta}_m,y)-\tilde l_m(\vz_m,y)\vert\le \sqrt{2}r\Vert \mW_m\Vert.
\end{align}

2. (Second Order Property) Smaller $\Vert \mW_m\Vert_F$ leads to smaller $\mu_m$ and $\beta_m$.
\begin{gather}
    \sum_i\lambda_i(\mH_m)=\text{tr}(\mH_m)=\text{tr}(\mW_m\mJ_m\mW_m^T)=\text{tr}(\mW_m^T\mW_m\mJ_m)\nonumber\\
    \le \Vert\mW_m\Vert_F^2(\sum_j p_{m,j}-p_{m,j}^2),
\end{gather}
where $\lambda_i(\mH_m)$ means the eigenvalues of $\mH_m$ in increasing order. $\lambda_1(\mH_m)=\mu_m$ and $\lambda_{-1}(\mH_m)=\beta_m$.

We notice that when increasing $\lambda_m$, namely, decreasing $\Vert \mW_m\Vert$ and $\Vert \mW_m\Vert_F$, we will decrease $g_m,\tilde c_m,\mu_m$ and $\beta_m$. According to \cref{theorem:robustness}, smaller $g_m$ will lead to stronger robustness while smaller $\mu_m$ will lead to weaker robustness. And according to \cref{theorem:inconsistency}, smaller $\tilde c_m$ and $\beta_m$ can lead to smaller objective inconsistency and thus better natural accuracy.

\section{Experiment Settings and Details}\label{Apx:exp_detail}
We run all the experiments on a sever with a single NVIDIA TITAN RTX GPU and an Intel Xeon Gold 6254 CPU.

\subsection{Details of Baselines}

\paragraph{FedDynAT~\cite{shah2021adversarial}.} FedDynAT proposes to use an annealing number of local training iterations to alleviate the slow convergence issue of Federated Adversarial Training (FAT)~\cite{zizzo2020fat}. More specifically, they anneal the number of local training iterations as $\tau_t = \tau_0\gamma_E^{t/F_E}$ where $\tau_t$ is the number of local training iterations at round $t$, $\gamma_E$ is the decay rate and $F_E$ is the decay period. When implementing FedDynAT, we use FedNOVA instead of FedCurv~\cite{shoham2019overcoming} to avoid extra communication in our resource-constrained settings. 

\paragraph{FedRBN~\cite{hong2021federated}.} FedRBN adopts Dual Batch Normalization (DBN) layers~\cite{xie2020adversarial} with two sets of batch normalization (BN) statistics for clean examples and adversarial examples respectively. When propagating the robustness from the clients who perform AT to the clients who perform ST, they use the adversarial BN statistics of AT clients to evaluate the adversarial BN statistics of ST clients as follows:
\begin{align}
    \mu_{\text{ST}}^a &= \mu_{AT}^a+\lambda_{\text{RBN}}(\mu_{\text{ST}}^n-\mu_{\text{AT}}^n),\\
    ({\sigma_{\text{ST}}^a})^2 &= ({\sigma_{\text{AT}}^a})^2\left[\frac{({\sigma_{\text{ST}}^n})^2}{({\sigma_{\text{AT}}^n})^2+\epsilon}\right]^{\lambda_{\text{RBN}}}
\end{align}
where $\mu_{\text{ST}}^a$ and $\mu_{\text{ST}}^n$ are the means in adversarial BN and natural BN respectively on a ST client, and $(\sigma_{\text{ST}}^a)^2$ and $(\sigma_{\text{ST}}^n)^2$ are the variances. Similarly, for AT clients we have $\mu_{\text{AT}}^a$, $\mu_{\text{AT}}^n$, $(\sigma_{\text{AT}}^a)^2$ and $(\sigma_{\text{AT}}^n)^2$. $\lambda_{\text{RBN}}$ is the hyperparameter and $\epsilon$ is a small constant. With these evaluations of the adversarial BN statistics, the ST clients can also attain some adversarial robustness without performing AT.

\subsection{Hyperparameters}
\paragraph{Hyperparameters of FL} To simulate the statistical heterogeneity in FL, we partition the whole dataset onto $N=100$ clients with the same Non-IID data partition as \citet{shah2021adversarial}, where $80\%$ data of each client is from only two classes while $20\%$ is from the other eight classes. We sample $30$ clients for local training in each communication round. For global FL (FedNOVA), the validation sets on all clients are I.I.D.. For personalized FL (FedBN), we make the validation set on each client have the same distribution as the training set on that client (i.e., Non-I.I.D.). We report the averaged validation accuracy over all clients in our experiments.

We set the number of initial local training iterations as $\tau_0=40$ for both FMNIST and CIFAR-10, and the local batch size is set to be $B=50$. We use the same trick as \citet{shah2021adversarial} that we gradually decrease the number of local training iterations. When training with FedNOVA, the maximal number of communication rounds is set to be $T=500$ for FMNIST and $T=2000$ for CIFAR-10. When training with FedBN, we set $T=150$ for FMNIST and $T=500$ for CIFAR-10. We use the SGD optimizer with a constant learning rate $\eta=0.01$ and momentum $0.9$ in all the experiments.

\paragraph{Hyperparameters of AT} Following \citet{moosavi2019robustness} and \citet{zizzo2020fat}, we adopt $\epsilon_0=0.15$ and $\alpha_0=0.03$ for FMNIST, and we use $\epsilon_0=\sfrac{8}{255}$ and $\alpha_0=\sfrac{2}{255}$ for CIFAR-10. We use PGD with $10$ iterations for training and testing in all our experiments. Same as \citet{zizzo2020fat}, we use a warmup phase with only standard training before performing any AT in all the experiments. For FedNOVA, the length of the warmup phase is set to be $50$ for FMNIST and $400$ for CIFAR-10. For FedBN, the length of the warmup phase is set to be $15$ for FMNIST and $200$ for CIFAR-10.

\paragraph{Hyperparameters of FADE} \cref{table:hyperparameters} summarizes the $\lambda_m$ and $\epsilon_{m-1}$ that we used in the experiments in \cref{Subsec:performance}. When tuning both $\epsilon_m$ and $\lambda_m$, we adopt the overall accuracy on both clean and adversarial examples as the criterion, which can be written as $A = 0.4A_n+0.6A_a$ where $A_n$ and $A_a$ are natural accuracy and adversarial accuracy respectively. When determining the feature perturbation $\epsilon_m$, we perform a linear search for the optimal discount factors $d_m\in[0.0,1.0]$ such that $\epsilon_m=d_m\epsilon_0$ for each $m\in[1,M]$. When determining the weight decay hyperparameter $\lambda_m$, we use the same $\lambda_m=\lambda$ for all modules, and we select the optimal $\lambda\in\{0.0001,0.0003,0.001,0.003,0.01,0.03,0.1\}$. We also show the model architectures and model partitions used in our experiments in \cref{table:cnn7} and \cref{table:vgg11}.

\paragraph{Hyperparameters of FedDynAT} We set the decay rate $\gamma_E=0.9$, and the decay period $F_E=T/10$.

\paragraph{Hyperparameters of FedRBN} Following \citet{hong2021federated}, we adopt the same setting where $\lambda_{\text{RBN}}=0.1$. We loose the requirement of a noise detector and allow an optimal noise detector for FedRBN such that it can always use the correct BN statistics during test (This makes its robustness stronger than that with a real noise detector).



\begin{table}[t]
\centering
\caption{The hyperparameters $\epsilon_m$ and $\lambda$ of FADE used in \cref{Subsec:performance}. }
\label{table:hyperparameters}
\begin{tabular}{cc|cc|cc|cc|cc|c}
\hline
\multirow{2}{*}{Model}  & \multirow{2}{*}{Optimizer} & \multicolumn{2}{c|}{Module 1} & \multicolumn{2}{c|}{Module 2} & \multicolumn{2}{c|}{Module 3} & \multicolumn{2}{c|}{Module 4} & \multirow{2}{*}{$\lambda$}\\ \cline{3-10} 
                        &                            & $\epsilon_0$ & $\alpha_0$     & $\epsilon_1$ & $\alpha_1$     & $\epsilon_2$ & $\alpha_2$    & $\epsilon_3$ & $\alpha_3$    &                            \\ \hline
\multirow{2}{*}{\begin{tabular}[c]{@{}c@{}}2-module CNN-7\\ FMNIST\end{tabular}}    & FedNOVA & 0.15 & 0.03 & 0.06 & 0.012 & n/a       & n/a           & n/a          & n/a           & 0.003                   \\ \cline{2-11} 
                                                                                    & FedBN   & 0.15 & 0.03 & 0.12 &0.024  & n/a       & n/a            & n/a         & n/a           & 0.03                     \\ \hline
\multirow{2}{*}{\begin{tabular}[c]{@{}c@{}}2-module VGG-11\\ CIFAR-10\end{tabular}} & FedNOVA & $\sfrac{8}{255}$ & $\sfrac{2}{255}$ & $\sfrac{3}{255}$ & $\sfrac{0.75}{255}$ & n/a & n/a & n/a & n/a & 0.1      \\ \cline{2-11}
                                                                                    & FedBN   & $\sfrac{8}{255}$ & $\sfrac{2}{255}$ & $\sfrac{3}{255}$ & $\sfrac{0.75}{255}$ & n/a & n/a & n/a & n/a & 0.003     \\ \hline
\multirow{2}{*}{\begin{tabular}[c]{@{}c@{}}3-module VGG-11\\ CIFAR-10\end{tabular}} & FedNOVA & $\sfrac{8}{255}$ & $\sfrac{2}{255}$ & $\sfrac{4}{255}$ & $\sfrac{1}{255}$ & $\sfrac{3}{255}$ & $\sfrac{0.75}{255}$ & n/a & n/a                                                                                      & 0.003        \\ \cline{2-11} 
                                                                                    & FedBN   & $\sfrac{8}{255}$ & $\sfrac{2}{255}$ & $\sfrac{4}{255}$ & $\sfrac{1}{255}$ & $\sfrac{3}{255}$ & $\sfrac{0.75}{255}$ & n/a & n/a & 0.001         \\ \hline
\multirow{2}{*}{\begin{tabular}[c]{@{}c@{}}4-module VGG-11\\ CIFAR-10\end{tabular}} & FedNOVA & $\sfrac{8}{255}$ & $\sfrac{2}{255}$ & $\sfrac{4}{255}$ & $\sfrac{1}{255}$ & $\sfrac{3}{255}$ & $\sfrac{0.75}{255}$ & $\sfrac{3}{255}$ & $\sfrac{0.75}{255}$                                                        & 0.03         \\ \cline{2-11} 
                                                                                    & FedBN   & $\sfrac{8}{255}$ & $\sfrac{2}{255}$ & $\sfrac{4}{255}$ & $\sfrac{1}{255}$  & $\sfrac{3}{255}$ & $\sfrac{0.75}{255}$ & $\sfrac{3}{255}$ & $\sfrac{0.75}{255}$ & 0.001         \\ \hline
\end{tabular}
\end{table}

\begin{table}[t]
\centering
\caption{The model architecture of CNN-7 and the model partitions of FADE with 1 module and 2 modules. We show the number of parameters of each module in the table.}
\label{table:cnn7}
\begin{tabular}{|c|c|c|c|}
\hline
Layer & Details                                                                                                                                             & 1 Module                  & 2 Modules                                                                                \\ \hline
1     & \begin{tabular}[c]{@{}c@{}}Conv2D (8, kernel size = 3, padding = 1, stride = 1)\\ BN2D, ReLU, MaxPool2D (kernel size = 2, stride = 2)\end{tabular}  & \multirow{7}{*}{38.874k} & \multirow{4}{*}{\begin{tabular}[c]{@{}c@{}}15.312k\\ + 2.89k\\ = 18.202k\end{tabular}} \\ \cline{1-2}
2     & \begin{tabular}[c]{@{}c@{}}Conv2D (16, kernel size = 3, padding = 1, stride = 1)\\ BN2D, ReLU, MaxPool2D (kernel size = 2, stride = 2)\end{tabular} &                           &                                                                                          \\ \cline{1-2}
3     & \begin{tabular}[c]{@{}c@{}}Conv2D (32, kernel size = 3, padding = 1, stride = 1)\\ BN2D, ReLU\end{tabular}                                          &                           &                                                                                          \\ \cline{1-2}
4     & \begin{tabular}[c]{@{}c@{}}Conv2D (32, kernel size = 3, padding = 1, stride = 1)\\ BN2D, ReLU, MaxPool2D (kernel size = 2, stride = 2)\end{tabular} &                           &                                                                                          \\ \cline{1-2} \cline{4-4} 
5     & \begin{tabular}[c]{@{}c@{}}Conv2D (64, kernel size = 3, padding = 0, stride = 1)\\ BN2D, ReLU\end{tabular}                                          &                           & \multirow{3}{*}{23.562k}                                                                \\ \cline{1-2}
6     & FC (64, 64), BN1D, ReLU                                                                                                                             &                           &                                                                                          \\ \cline{1-2}
7     & FC (64, 10)                                                                                                                                         &                           &                                                                                          \\ \hline
\end{tabular}
\end{table}

\begin{table}[t]
\centering
\caption{The model architecture of VGG-11 and the model partitions of FADE with 1 module, 2 modules, 3 modules and 4 modules. We show the number of parameters of each module (backbone layers + auxiliary model) in the table.}
\label{table:vgg11}
\begin{tabular}{|c|c|c|c|c|c|}
\hline
Layer & Details                                                                                                                                              & 1 Module                 & 2 Modules                                                                             & 3 Modules                                                                             & 4 Modules                                                                             \\ \hline
1     & \begin{tabular}[c]{@{}c@{}}Conv2D (64, kernel size = 3, padding = 1, stride = 1)\\ BN2D, ReLU, MaxPool2D (kernel size = 2, stride = 2)\end{tabular}  & \multirow{11}{*}{9.758M} & \multirow{6}{*}{\begin{tabular}[c]{@{}c@{}}4.504M\\ + 0.005M\\ = 4.509M\end{tabular}} & \multirow{4}{*}{\begin{tabular}[c]{@{}c@{}}0.962M\\ + 0.010M\\ = 0.972M\end{tabular}} & \multirow{5}{*}{\begin{tabular}[c]{@{}c@{}}2.143M\\ + 0.020M\\ = 2.163M\end{tabular}} \\ \cline{1-2}
2     & \begin{tabular}[c]{@{}c@{}}Conv2D (128, kernel size = 3, padding = 1, stride = 1)\\ BN2D, ReLU, MaxPool2D (kernel size = 2, stride = 2)\end{tabular} &                          &                                                                                       &                                                                                       &                                                                                       \\ \cline{1-2}
3     & \begin{tabular}[c]{@{}c@{}}Conv2D (256, kernel size = 3, padding = 1, stride = 1)\\ BN2D, ReLU\end{tabular}                                          &                          &                                                                                       &                                                                                       &                                                                                       \\ \cline{1-2}
4     & \begin{tabular}[c]{@{}c@{}}Conv2D (256, kernel size = 3, padding = 1, stride = 1)\\ BN2D, ReLU, MaxPool2D (kernel size = 2, stride = 2)\end{tabular} &                          &                                                                                       &                                                                                       &                                                                                       \\ \cline{1-2} \cline{5-5}
5     & \begin{tabular}[c]{@{}c@{}}Conv2D (512, kernel size = 3, padding = 1, stride = 1)\\ BN2D, ReLU\end{tabular}                                          &                          &                                                                                       & \multirow{2}{*}{\begin{tabular}[c]{@{}c@{}}3.542M\\ + 0.005M\\ = 3.547M\end{tabular}} &                                                                                       \\ \cline{1-2} \cline{6-6} 
6     & \begin{tabular}[c]{@{}c@{}}Conv2D (512, kernel size = 3, padding = 1, stride = 1)\\ BN2D, ReLU, MaxPool2D (kernel size = 2, stride = 2)\end{tabular} &                          &                                                                                       &                                                                                       & \begin{tabular}[c]{@{}c@{}}2.361M\\ + 0.005M\\ = 2.366M\end{tabular}                  \\ \cline{1-2} \cline{4-6} 
7     & \begin{tabular}[c]{@{}c@{}}Conv2D (512, kernel size = 3, padding = 1, stride = 1)\\ BN2D, ReLU\end{tabular}                                          &                          & \multirow{5}{*}{5.254M}                                                               & \multirow{5}{*}{5.254M}                                                               & \begin{tabular}[c]{@{}c@{}}2.361M\\ + 0.005M\\ = 2.366M\end{tabular}                  \\ \cline{1-2} \cline{6-6} 
8     & \begin{tabular}[c]{@{}c@{}}Conv2D (512, kernel size = 3, padding = 1, stride = 1)\\ BN2D, ReLU, MaxPool2D (kernel size = 2, stride = 2)\end{tabular} &                          &                                                                                       &                                                                                       & \multirow{4}{*}{2.893M}                                                               \\ \cline{1-2}
9     & FC (512, 512), BN1D, ReLU                                                                                                                            &                          &                                                                                       &                                                                                       &                                                                                       \\ \cline{1-2}
10    & FC (512, 512), BN1D, ReLU                                                                                                                            &                          &                                                                                       &                                                                                       &                                                                                       \\ \cline{1-2}
11    & FC (512, 10)                                                                                                                                         &                          &                                                                                       &                                                                                       &                                                                                       \\ \hline
\end{tabular}
\end{table}


\end{document}